\documentclass[final]{cvpr}

\usepackage{times}
\usepackage{epsfig}
\usepackage{graphicx}
\usepackage{amsmath}
\usepackage{amssymb}
\usepackage{algorithm}
\usepackage{algorithmic}
\usepackage{bm}
\usepackage{cite}
\usepackage{multirow}
\usepackage{makecell}
\newcommand{\PreserveBackslash}[1]{\let\temp=\\#1\let\\=\temp}
\newcolumntype{C}[1]{>{\PreserveBackslash\centering}p{#1}}
\newcolumntype{R}[1]{>{\PreserveBackslash\raggedleft}p{#1}}
\newcolumntype{L}[1]{>{\PreserveBackslash\raggedright}p{#1}}
\usepackage[pagebackref=true,breaklinks=true,colorlinks,bookmarks=false]{hyperref}
\def\cvprPaperID{1798} 
\def\confYear{CVPR 2021}

\begin{document}

\title{From Rain Generation to Rain Removal}
\vspace{-3mm}
\author{Hong Wang$^{1,2,}$\footnotemark[1], Zongsheng Yue$^{1,}$\footnotemark[1], Qi Xie$^{1}$, Qian Zhao$^{1}$, Yefeng Zheng$^{2}$, Deyu Meng$^{1,}$\footnotemark[2]\\
$^{1}$Xi'an Jiaotong University, Xi'an, China \quad
$^{2}$Tencent Jarvis Lab, Shenzhen, China\\
{\tt\small dymeng@mail.xjtu.edu.cn}
}
\maketitle
\renewcommand{\thefootnote}{\fnsymbol{footnote}}
\footnotetext[2]{Corresponding author}
\renewcommand{\thefootnote}{\arabic{footnote}}
\renewcommand{\thefootnote}{\fnsymbol{footnote}}
\footnotetext[1]{Equal contribution}
\renewcommand{\thefootnote}{\arabic{footnote}}
\thispagestyle{empty}

\begin{abstract}
For the single image rain removal (SIRR) task, the performance of deep learning (DL)-based methods is mainly affected by the designed deraining models and training datasets. Most of current state-of-the-art focus on constructing powerful deep models to obtain better deraining results. In this paper, to further improve the deraining performance, we novelly attempt to handle the SIRR task from the perspective of training datasets by exploring a more efficient way to synthesize rainy images. Specifically, we build a full Bayesian generative model for rainy image where the rain layer is parameterized as a generator with the input as some latent variables representing the physical structural rain factors, e.g., direction, scale, and thickness. To solve this model, we employ the variational inference framework to approximate the expected statistical distribution of rainy image in a data-driven manner. With the learned generator, we can automatically and sufficiently generate diverse and non-repetitive training pairs so as to efficiently enrich and augment the existing benchmark datasets. User study qualitatively and quantitatively evaluates the realism of generated rainy images. Comprehensive experiments substantiate that the proposed model can faithfully extract the complex rain distribution that not only helps significantly improve the deraining performance of current deep single image derainers, but also largely loosens the requirement of large training sample pre-collection for the SIRR task.
\end{abstract}

\begin{figure}[t]
  \centering\vspace{-3mm}
  \includegraphics[width=0.9\linewidth]{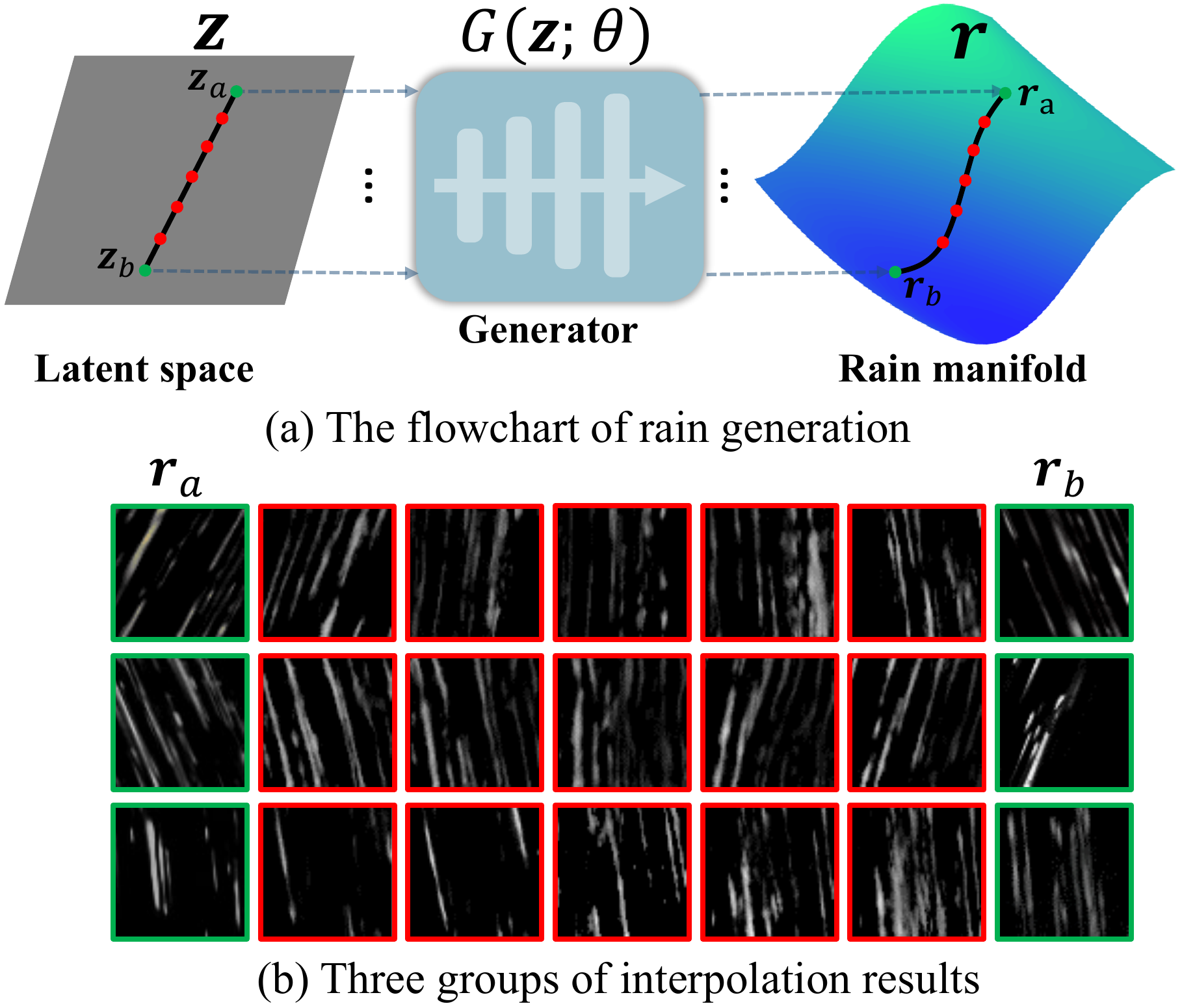} 
    \caption{Interpolation results in latent space $\boldsymbol{z}$ representing rain factors. (a) The rain distribution is implicitly modeled as a generator $G$; (b) Three groups of generated rain layers through interpolations in latent space. For each group, $\boldsymbol{r}_a$ and $\boldsymbol{r}_b$ (marked as green) represent the rain layers in the original training dataset, while the ones (marked as red) between them are generated from latent codes (marked as red points) in $\boldsymbol{z}$ space. These codes are obtained by linearly interpolating between $\boldsymbol{z}_a$ and $\boldsymbol{z}_b$ which are the latent codes of $\boldsymbol{r}_a$ and $\boldsymbol{r}_b$, respectively.}
  \label{intro}
  \vspace{-4mm}
\end{figure}
\vspace{-1mm}\section{Introduction}\vspace{-1mm}
Recently, single image rain removal (SIRR) has attracted considerable attention, which is usually regarded as a necessary pre-processing step of outdoor image processing tasks, e.g., autonomous driving~\cite{geiger2012are}, scene segmentation~\cite{cheng2012outdoor}, and object tracking~\cite{Comaniciu2003Kernel}. Due to the complex and diverse rain structures in real scenes, SIRR is still a typical challenging task in computer vision~\cite{benchmark,Yang2020Single}.

Driven by massive training data (rainy images) and the powerful fitting capability of deep convolutional neural network (CNN), deep learning (DL) represents the current research trend in the SIRR task. Clearly, the performance of DL-based methods is mainly affected by two key factors, i.e., the rationality and capacity of deraining models and the quality of training datasets. Most of current works focus on the former and aim to improve the deraining results mainly by building more sophisticated networks~\cite{Fu2017Clearing,qian,zhang2018density,wang2019spatial,Fu2017Removing,Li2018Non,wang2019erl,hu2019depth,du2020variational,li2018recurrent,ren2019progressive,Yang2019Joint,yasarla2019uncertainty,jiang2020multi} and designing better learning manners~\cite{li2019heavy, Mu2019Learning,wang2020model,wei2019semi,yasarla2020syn2real,jin2019unsupervised}. Albeit achieving satisfied performance in some scenarios, they put less emphasis on the impact of training data and largely rely on the off-the-shelf datasets to train their deraining models. Curiously, are the existing datasets sufficiently good?  Is it possible to further improve the performance of current DL-based derainers directly by ameliorating the quality of these datasets? This paper mainly concentrates on these issues.

Currently, for the SIRR task, the existing datasets are mainly obtained by the following manners: 1) The common one is to synthesize rain streaks with the photo-realistic rendering technique~\cite{garg2006photorealistic} and then add them on clear images~\cite{Yang2019Joint, Fu2017Removing,Li2016Rain,zhang2019image,zhang2018density,benchmark}. 2) Instead of such simple addition operation, inspired by ~\cite{gk_vision}, some works~\cite{halder2019physics,hu2019depth} explored better fusion mechanisms between clear images and rain streaks. However, the exploited rains are still synthesized by manually setting some oscillation parameters of raindrops~\cite{garg2006photorealistic}. 3) The unpaired image translation strategy is another new generation manner, which attempts to learn a mapping from a clean image to rainy one with adversarial learning  so as to generate paired rainy-clean images, such as \cite{wei2019deraincyclegan,zhu2019singe,pizzati2020model}. 4) There is one real rain dataset proposed by ~\cite{wang2019spatial}, which is semi-automatically generated through rain videos shot in real rain scenes by manually adjusting camera parameters, including exposure duration and ISO.

Although these existing datasets can be used to train deep derainers to some extent, their generation manners still possess some evident limitations. Specifically, for 1) and 2), rains are synthesized by empirically setting some parameters through human subjective assumptions, which would restrict the generated rain types. Besides, the acquisition process of training samples needs human supervision and physical simulators. This is time-consuming and labor-cumbersome. As for 3), the intrinsic mechanisms of rains are more or less ignored and thus it has less physical interpretability. While for 4), it is always hard to shoot enough rain scenes for sufficiently representing the complicated rain shapes in real world. All these deficiencies tend to adversely affect the quality and the diversity of training datasets and limit the performance improvement of current deep SIRR derainers. Thus, it is critical to build a proper model representing the rain statistical distribution in order to automatically  and faithfully generate diverse rains.


In this work, we attempt to explore the intrinsic generative mechanism underlying rain streaks and propose a better generation process. As seen in Fig.~\ref{intro}, high quality of rain streaks, with diverse and non-repetitive shapes, can be easily obtained through the learned generator $G$ by our method that represents the implicit distribution of rains. It is worth mentioning that the generated rain streaks $\boldsymbol{r}$  (marked as red in Fig.~\ref{intro}(b)) exhibit more unseen patterns in the original training dataset $\boldsymbol{r}_a$ and $\boldsymbol{r}_b$ (marked as green).
Especially, such generator with an explicit mapping form tends to \mbox{provide} intrinsic clues for understanding the generation of rains, which is meaningful for general tasks
on rainy images. In summary, our contributions are mainly three-fold:

Firstly, this work specifically proposes a generative model to depict the generation process of a rainy image. Specifically, different from hand-crafted priors for rains~\cite{Yu2015Removing,Gu2017Joint,wei2019semi} or physics-based imaging analysis about rains~\cite{gk_vision}, the proposed model makes effort to explore an implicit distribution of rain layer in statistics. A deep variational inference algorithm is specifically designed to to approximate the expected distribution of rainy images.

Secondly, an interpretable rain generator can be obtained, capable of delivering the intrinsic manifold projection from latent factors, such as direction and thickness, to rain streaks as shown in Fig.~\ref{intro}. This makes it possible to efficiently generate diverse and non-repetitive rain streaks without subjective human intervention and empirical parameter settings. Disentanglement and interpolation experiments substantiate the rationality of the proposed generator, and a user study evaluates the realism of generated rainy images. Moreover, the small sample experiment exhibits the potentials of the proposed model in real applications.

Thirdly, the proposed generator facilitates an easy augmentation of diverse rains for current DL-based SIRR derainers. Comprehensive experiments on synthetic and real datasets validate that the performance of these DL-based derainers can be significantly improved by retraining them on augmented datasets. This coincides with our motivation that improving the quality of datasets is rational and helpful.

\vspace{-1mm}\section{Related Work}\label{works}\vspace{0mm}

\textbf{Rain Dataset Synthesizing.} Previously, Garg and Nayar analyzed the appearance and imaging process of rain~\cite{gk_vision} and synthesized a rain streak database with the photo-realistic rendering technique~\cite{garg2006photorealistic}. Similarly, researchers synthesize different rain streaks and then add them on clear images to construct paired samples such as Rain100H~\cite{Yang2019Joint}, Rain1400~\cite{Fu2017Removing}, Rain800~\cite{zhang2019image}, and DID-MDN~\cite{zhang2018density}. Besides, there are some works exploring how to merge rains with background images, for example, RainCityscapes~\cite{hu2019depth}, NYU-Rain~\cite{li2019heavy}, and MPID~\cite{benchmark}.
Recently, the unpaired image translation idea is widely adopted to generate weather-corrupted images~\cite{wei2019deraincyclegan,zhu2019singe,pizzati2020model}. For example,
Pizzati \emph{et al.}~\cite{pizzati2020model} proposed to disentangle the scene from occlusions, e.g., raindrops and dirt, which can generate realistic translations. These generation methods largely rely on human subjective assumptions and often require setting model parameters, which would limit the diversity of synthesized rains.

Instead of synthesizing rains, Wang \emph{et al.}~\cite{wang2019spatial} proposed a large-scale real rain dataset, called SPA-Data, which was semi-automatically generated from real rain videos shot in real rain scenes or collected from Internet. To \mbox{construct} paired samples, the clean images are roughly estimated based on successive several frames. The main limitation of this dataset is that the expensive cost of shooting rain scenes makes it difficult to capture large number of rain types.


\textbf{Rain Removal.}
Very recently, for the SIRR task, researchers have designed various network structures, from simple CNN~\cite{Fu2017Clearing,Fu2017Removing} to complicated recurrent and multi-stage learning~\cite{yang2017deep,Yang2019Joint,li2018recurrent,ren2019progressive}.
Besides, some works incorporate multi-scale learning to exploit the self-similarity both within the same scale or across different scales~\cite{fu2019lightweight,zheng2019residual,yasarla2019uncertainty,jiang2020multi}. There are also some other network frameworks, for example, adversarial learning~\cite{zhang2018density,zhang2019image, wei2019deraincyclegan,li2019heavy,Li2020All}, encoder-decoder ~\cite{Li2018Non,wang2019erl,hu2019depth,du2020variational}, and semi-/un-supervised learning~\cite{wei2019semi,yasarla2020syn2real,wei2019deraincyclegan,jin2019unsupervised,zhu2019singe}. Now there is another novel research line that prior knowledge is embedded into deep networks to improve the interpretability, such as~\cite{Mu2019Learning,wei2019semi,wang2020model}.

Although these DL-based techniques have achieved remarkable success, they mainly utilize the aforementioned off-the-shelf datasets as training data. In this work, we aim to explore an automatic generative mechanism with the capability to simulate possibly variant rain types, for ameliorating the quality of the existing datasets and thus expectantly improving the deraining results of current deep derainers.

\textbf{Generative Models.}
As an active research topic in computer vision and machine learning, deep generative models have been widely studied recently, such as variational autoencoder (VAE)~\cite{kingma2013auto,rezende2014stochastic}, generative adversarial network (GAN)~\cite{goodfellow2014generative,radford2015unsupervised}, and flow-based generative model~\cite{dinh2014nice}. Especially, as prominent models, VAE and GAN have achieved remarkable success in many image generation tasks, including face modeling~\cite{kupyn2018deblurgan,li2018learning}, style transfer~\cite{zhu2017unpaired}, image noise generation~\cite{chen2018image,kim2019grdn} and so on. To the best of our knowledge, there is still little work completely focusing on the rain generation task. Therefore, inspired by these deep generative models, we take a step forward to explore the intrinsic generative mechanisms of rainy images as well as rain streaks.
\vspace{-2mm}\section{The Proposed Method}\vspace{-1mm}\label{section3}
Given a training set $\mathcal{D}=\left \{\boldsymbol{o}_{n},\boldsymbol{x}_{n}\right \}_{n=1}^{N}$, where $\boldsymbol{o}_{n}$ is the $n$-th rainy image and $\boldsymbol{x}_{n}$ is the background, we aim to explore the physical mechanism of the rainy image and learn its underlying distribution. To this aim, we \mbox{construct} a generative model for rainy image under the Bayesian framework by implicitly modeling rain layer as a generator.
With our specifically designed inference algorithm, the model can extract the general statistical distribution of rainy image as well as rain streak based on the training dataset in a data-driven manner. This enables the free generation of rains with diverse shapes. The details are given below.
\vspace{-1mm}\subsection{Generative Model}\vspace{-0mm}
Similar to~\cite{wang2019spatial,ren2019progressive,wang2020model}, given any single rainy image
$\boldsymbol{o}\in \mathbb{R}^{d}$ with size $d$ as height $\times$ width, the generation process is:
\vspace{-2mm}
\begin{equation}\label{o}
\boldsymbol{o}=\boldsymbol{r}+\boldsymbol{b},
\vspace{-1mm}
\end{equation}
where $\boldsymbol{r}$ and $\boldsymbol{b}$ denote the rain layer and latent clean background underlying $\boldsymbol{o}$, respectively. Therefore, the generation of rainy image decomposes into two parts as follows:

\textbf{Rain Modeling.} For rain layer $\boldsymbol{r}$, it is difficult to depict it by using an accurate distribution in statistics. But it is very intuitive that the appearance of rain can be represented by some evident latent factors, such as direction, scale, and thickness~\cite{zhang2018density,li2018video,wang2020model}. Motivated by this observation, we encode such physical structural factors underlying rains as latent variable $\boldsymbol{z}\in\mathbb{R}^{t}$ and
generally model the rain layer $\boldsymbol{r}$ as a deep generator conditioned on $\boldsymbol{z}$, i.e.,
\vspace{-2mm}
\begin{equation}\label{rsample}
\boldsymbol{r}=G(\boldsymbol{z}; \theta),
\vspace{-2mm}
\end{equation}
where $\theta$ denotes the  parameters of the generator $G$.

As suggested in~\cite{burgess2018understanding,kim2018disentangling}, the isotropic Gaussian prior distribution is imposed on $\boldsymbol{z}$ as:
\vspace{-2mm}
\begin{equation}\label{gaussz}
\boldsymbol{z}\sim \mathcal{N}\left (\boldsymbol{z}|\boldsymbol{0},\textbf{I}_{t}\right ),
\vspace{-2mm}
\end{equation}
where $\textbf{I}_{t}\in \mathbb{R}^{t\times t}$ is the unit matrix. Such a prior has the potential to disentangle the physical rain factors in $\boldsymbol{z}$. This is visually validated in Section~\ref{disen}.


\textbf{Background Modeling.} In the given training pairs, the rain-free image $\boldsymbol{x}$ is usually simulated or estimated based on multiple rainy images taken on the same condition like real SPA-Data~\cite{wang2019spatial}, and it is not the exact latent clean background $\boldsymbol{{b}}$. We thus embed $\boldsymbol{x}$ into the following Gaussian prior distribution to constrain $\boldsymbol{b}$ as:
\vspace{-2mm}
\begin{equation}\label{gaussb}
\boldsymbol{{b}}\sim \mathcal{N}\left (\boldsymbol{b}|\boldsymbol{x},\varepsilon_{0}^{2}\textbf{I}_{d}\right ),
\vspace{-2mm}
\end{equation}
where $\varepsilon_{0}^{2}$ is a hyper-parameter measuring the similarity between $\boldsymbol{x}$ and $\boldsymbol{b}$,
and can be easily set as a small value. Note that for synthetic data where rainy images are obtained by adding synthesized rains on the pre-collected
images~\cite{Li2016Rain,Fu2017Removing,Yang2019Joint}, $\boldsymbol{x}$ can be regarded as the true groundtruth $\boldsymbol{b}$. In this case, Dirac prior on $\boldsymbol{b}$ is a proper choice which can be well approximated by setting $\varepsilon_{0}^{2}$ close to 0 in Eq.~(\ref{gaussb}).

From Eqs.~(\ref{o})-(\ref{gaussb}), it is easy to derive a full Bayesian model. Specifically, for rainy image $\boldsymbol{o}$, the likelihood is:
\vspace{-2mm}
\begin{equation}\label{cond}
\boldsymbol{o} \sim p_\theta \left (\boldsymbol{o}|\boldsymbol{z},\boldsymbol{b} \right ).
\vspace{-2mm}
\end{equation}
Note that $p_\theta \left (\cdot\right)$ means that this implicit distribution also relies on the parameters $\theta$ of generator $G$ defined in Eq.~(\ref{rsample}).

Finally, the task of generating rainy image turns to learn the general statistical distribution $p\left ( \boldsymbol{o} \right )$, expressed as:\footnote{Here we assume that $\boldsymbol{z}$ and $\boldsymbol{b}$ are mutually independent.}
\vspace{-2mm}
\begin{equation}\label{marg}
\centering
p\left ( \boldsymbol{o} \right ) = \int\int p_\theta \left (\boldsymbol{o}|\boldsymbol{z},\boldsymbol{b} \right )p\left (\boldsymbol{z}\right )p\left (\boldsymbol{b} \right )d\boldsymbol{z}d\boldsymbol{b},
\vspace{-2mm}
\end{equation}
where $ p\left ( \boldsymbol{z} \right )$ and $ p\left ( \boldsymbol{b} \right )$ are the prior distributions of $\boldsymbol{z}$ and $\boldsymbol{b}$, corresponding to Eq.~(\ref{gaussz}) and Eq.~(\ref{gaussb}), respectively.

Since the integral in Eq.~(\ref{marg}) is intractable, next we adopt the variational Bayesian framework to learn the $ p\left ( \boldsymbol{o} \right )$.
\begin{figure*}[t]
  \centering
  \vspace{-3mm}
  \includegraphics[width=0.95\linewidth]{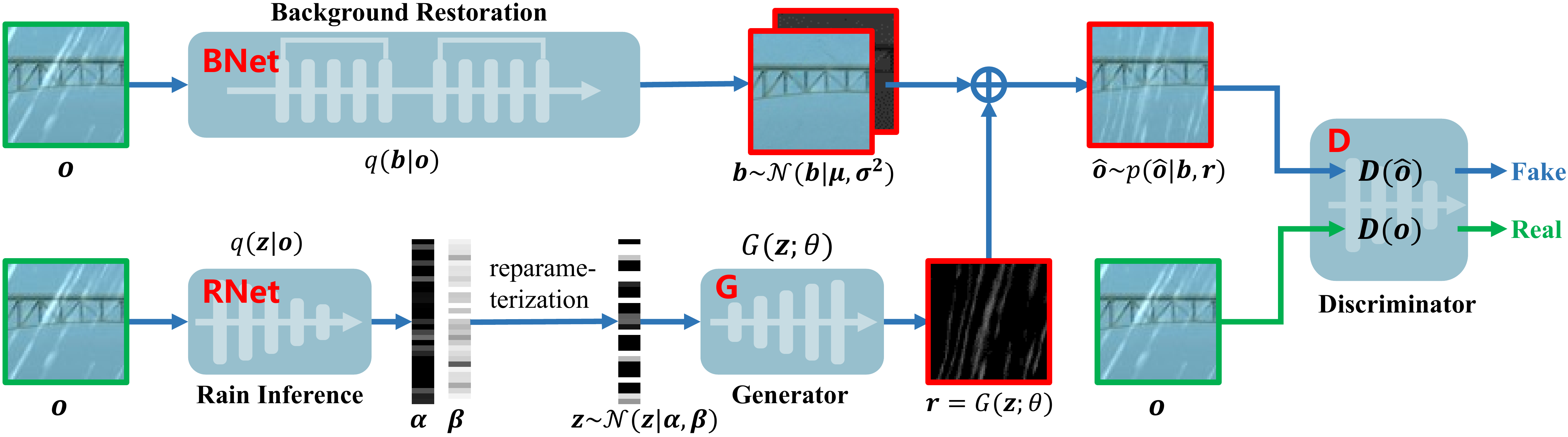}
  \caption{The flowchart of the proposed variational rain generation network (VRGNet). It contains four sub-networks, which are correspondingly constructed based on the estimation of $\textrm{log}\!\ p\left ( \boldsymbol{o} \right )$, i.e.,  ${\mathcal{L}}\left (\boldsymbol{z},\boldsymbol{b};\boldsymbol{o} \right )$ in Eq.~(\ref{elbo2}).}
  \vspace{-3mm}
  \label{net}
\end{figure*}
\vspace{-1mm}\subsection{Variational Object}
To learn $ p\left ( \boldsymbol{o} \right )$, we can decompose its logarithm as~\cite{blei2006variational}:\footnote{More derivations are included in supplementary material (SM).}
\vspace{-1mm}
\begin{equation}\label{hood}
\textrm{log}\!\ p\left ( \boldsymbol{o} \right ) \!=\!\mathcal{L}\left (\boldsymbol{z},\boldsymbol{b};\boldsymbol{o} \right )\!+\!D_{KL}\left [ q\left ( \boldsymbol{z},\boldsymbol{b}|\boldsymbol{o} \right ) || ~p\left (\boldsymbol{z},\boldsymbol{b}|\boldsymbol{o}\right )\right ],
\vspace{-1mm}
\end{equation}
where the first term in Eq.~(\ref{hood}) is expressed as:
\vspace{-1mm}
\begin{equation}\label{elbo}
\mathcal{L}\!\left (\boldsymbol{z},\boldsymbol{b};\boldsymbol{o} \right )\!=\!E_{q\left ( \boldsymbol{z},\boldsymbol{b}|\boldsymbol{o} \right )}\!\left [\textrm{log}\!\ p_{\theta}\!\left (\boldsymbol{o}|\boldsymbol{z},\boldsymbol{b} \right )\!p\!\left ( \boldsymbol{z} \right )\! p\!\left ( \boldsymbol{b}\right ) \!-\!\textrm{log}\!\ q\!\left(\boldsymbol{z},\boldsymbol{b}|\boldsymbol{o} \right )\right ].
\vspace{-1mm}
\end{equation}
Here $E_{p(a)}[f(a)]$ is the expectation of function $f(a)$ about the stochastic variable $a$ with the probability density function $p(a)$. $q\left ( \boldsymbol{z},\boldsymbol{b}|\boldsymbol{o} \right )$ is the variational approximate posterior of the true posterior $p\left (\boldsymbol{z},\boldsymbol{b}|\boldsymbol{o}\right )$ about the latent variables $\boldsymbol{z}$ and $\boldsymbol{b}$. The second term in Eq.~(\ref{hood}) is
the KL divergence measuring the difference between $q\left ( \boldsymbol{z},\boldsymbol{b}|\boldsymbol{o} \right )$ and $p\left (\boldsymbol{z},\boldsymbol{b}|\boldsymbol{o}\right )$. The non-negative property of KL divergence can thus lead to the following inequality, i.e.,
\vspace{-1mm}
\begin{equation}\label{inequ}
\textrm{log}\!\ p\left ( \boldsymbol{o} \right )\geq \mathcal{L}\left (\boldsymbol{z},\boldsymbol{b};\boldsymbol{o} \right ).
\vspace{-1mm}
\end{equation}
Thus the variational lower bound $\mathcal{L}\left (\boldsymbol{z},\boldsymbol{b};\boldsymbol{o} \right )$ can be viewed as an estimation of $\textrm{log}\!\ p\left ( \boldsymbol{o} \right )$ with error as the KL divergence. The learning of $p\left ( \boldsymbol{o} \right )$ can be achieved through approaching $\textrm{log}\!\ p\left ( \boldsymbol{o} \right )$ by maximizing its estimation  $\mathcal{L}\left (\boldsymbol{z},\boldsymbol{b};\boldsymbol{o} \right )$.

Based on the analysis above,
once $\theta$ is optimized by maximizing $\mathcal{L}\left (\boldsymbol{z},\boldsymbol{b};\boldsymbol{o} \right )$, the obtained explicit mapping $G(\boldsymbol{z}; \theta)$ can be directly used to synthesize rainy image, i.e., $\boldsymbol{o}=G(\boldsymbol{z}; \theta)+\boldsymbol{b}$, where $\boldsymbol{z}$ and $\boldsymbol{b}$ are sampled from $ p\left ( \boldsymbol{z} \right )$ and $ p\left ( \boldsymbol{b} \right )$, respectively. Therefore, for this generation task, the key problem is how to maximize the $\mathcal{L}\left (\boldsymbol{z},\boldsymbol{b};\boldsymbol{o} \right )$ in Eq.~(\ref{elbo}).
\subsection{Optimization}
Now we give the optimization algorithm for maximizing the $\mathcal{L}\left (\boldsymbol{z},\boldsymbol{b};\boldsymbol{o} \right )$. From Eq.~(\ref{elbo}), the key is to deal with the variational posterior $q\left (\boldsymbol{z},\boldsymbol{b}|\boldsymbol{o}\right )$ and the implicit $p_{\theta}\left (\boldsymbol{o}|\boldsymbol{z},\boldsymbol{b} \right )$.

As for $q\left (\boldsymbol{z},\boldsymbol{b}|\boldsymbol{o}\right )$, like the commonly-used factorized hypothesis in the mean-field variational inference~\cite{kingma2013auto}, we introduce the conditional independence assumption as:
\vspace{-2mm}
\begin{equation}\label{vp}
q\left (\boldsymbol{z},\boldsymbol{b}|\boldsymbol{o}\right ) = q\left (\boldsymbol{z}|\boldsymbol{o}\right )q\left (\boldsymbol{b}|\boldsymbol{o}\right ).
\vspace{-1mm}
\end{equation}
Then, the $\mathcal{L}\left (\boldsymbol{z},\boldsymbol{b};\boldsymbol{o} \right )$ in Eq.~(\ref{elbo}) can be equally rewritten as:
 \vspace{-1mm}
\begin{equation}\label{elbo2}
\begin{split}
\mathcal{L}\left (\boldsymbol{z},\boldsymbol{b};\boldsymbol{o} \right )&\!=E_{q\left ( \boldsymbol{z},\boldsymbol{b}|\boldsymbol{o} \right )}\left [\textrm{log}\!\ p_{\theta}\left (\boldsymbol{o}|\boldsymbol{z},\boldsymbol{b} \right )\right ]\\&\!\!\!\!\!\!\!\!\!\!\!\!\!\!\!\!\!\!-\!D_{KL}\left [q\left ( \boldsymbol{z}|\boldsymbol{o}\right )||~p\left ( \boldsymbol{z} \right )\right ]\!-\!D_{KL}\left [q\left ( \boldsymbol{b}|\boldsymbol{o}\right )||~p\left ( \boldsymbol{b} \right )\right ].
\end{split}
 \vspace{-1mm}
\end{equation}

To maximize $\mathcal{L}\left (\boldsymbol{z},\boldsymbol{b};\boldsymbol{o} \right )$ in Eq.~(\ref{elbo2}), we impose Gaussian distribution on the posteriors $q\left (\boldsymbol{z}|\boldsymbol{o}\right )$ and $q\left (\boldsymbol{b}|\boldsymbol{o}\right )$ to approach the Gaussian priors $p\left (\boldsymbol{z}\right )$ and $p\left (\boldsymbol{b}\right )$, respectively, i.e.,
\vspace{-2mm}
\begin{equation}\label{vpr}
 q\left (\boldsymbol{z}|\boldsymbol{o}\right ) =  \prod\nolimits_{i=1}^{t}\mathcal{N}\left ( z_{i}|\alpha_{i}\left (\boldsymbol{o};W_{R}\right ), \beta _{i}\left ( \boldsymbol{o};W_{R}\right )\right ),
 \vspace{-2mm}
\end{equation}
\vspace{-0mm}
\begin{equation}\label{vpb}
q\left (\boldsymbol{b}|\boldsymbol{o}\right ) = \prod\nolimits_{j=1}^{d}\mathcal{N}\left ( b_{j}|\mu_{j}\left (\boldsymbol{o};W_{B}\right ), \sigma _{j}^{2}\left ( \boldsymbol{o};W_{B}\right )\right ),
\vspace{-1mm}
\end{equation}
where $\alpha_{i}\left (\boldsymbol{o};W_{R}\right )$ and $\beta_{i}\left ( \boldsymbol{o};W_{R}\right )$ are functions for inferring the posterior parameters (i.e., mean and variance, respectively) of latent
variable $\boldsymbol{z}$, and they are integrally parameterized as one rain inference network, called \textit{RNet} with parameter $W_{R}$. $\mu_{j}\left (\boldsymbol{o};W_{B}\right )$ and $\sigma _{j}^{2}\left ( \boldsymbol{o};W_{B}\right )$ are functions from $\boldsymbol{o}$ to variational
posterior parameters of latent variable $\boldsymbol{b}$. They are jointly parameterized as another network, called \textit{BNet} with parameter $W_{B}$ for restoring clean background.

From Eqs.~(\ref{gaussz}), (\ref{gaussb}), (\ref{vpr}), and (\ref{vpb}), it is easy to compute the last two terms in Eq.~(\ref{elbo2}) as:
\small
\vspace{-1mm}
\begin{equation}\label{term}
\begin{split}
&D_{KL}\!\left [q\left ( \boldsymbol{z}|\boldsymbol{o}\right )||~p\left ( \boldsymbol{z} \right )\right]=\sum\nolimits_{i=1}^{t}\left \{ \frac{\alpha_{i}^{2}}{2} +\frac{1}{2}\left (\beta_{i} -\textrm{log}\beta _{i}-1\right )\right \},\\
&D_{KL}\!\left [q\left ( \boldsymbol{b}|\boldsymbol{o}\right )||~p\left ( \boldsymbol{b} \right )\right]\!=\!\sum\nolimits_{j=1}^{d}\!\!\left \{ \!\frac{\left (\mu_{j}- x_{j}\right )^{2}}{2\varepsilon_{0} ^{2}} \!+\!\frac{1}{2}\!\left (\! \frac{\sigma _{j}^{2}}{\varepsilon_{0} ^{2}} \!-\!\textrm{log}\frac{\sigma _{j}^{2}}{\varepsilon_{0} ^{2}}\!-\!1\!\right )\!\right \}.\\
\end{split}
\vspace{-9mm}
\end{equation}
\normalsize
where we simplify $\alpha_{i}\left (\boldsymbol{o};W_{R}\right )$, $\beta_{i}\left ( \boldsymbol{o};W_{R}\right )$, $\mu_{j}\left (\boldsymbol{o};W_{B}\right )$, and $\sigma _{j}^{2}\left ( \boldsymbol{o};W_{B}\right )$, as $\alpha_{i}$, $\beta_{i}$, $\mu_{j}$, and $\sigma _{j}^{2}$, respectively.

However, we cannot directly calculate the first term in Eq.~(\ref{elbo2}) due to the implicity of $p_{\theta}\left (\boldsymbol{o}|\boldsymbol{z},\boldsymbol{b} \right )$. Fortunately, the generator $G$ enables the sampling from $p_{\theta}\left (\boldsymbol{o}|\boldsymbol{z},\boldsymbol{b} \right )$, i.e.,
\vspace{-2mm}
\begin{equation}\label{imo}
\boldsymbol{o} \sim p_{\theta}\left (\boldsymbol{o}|\boldsymbol{z},\boldsymbol{b} \right )\Longleftrightarrow \boldsymbol{o}=G(\boldsymbol{z}; \theta)+\boldsymbol{b},
\vspace{-2mm}
\end{equation}
which motivates us to introduce a \emph{discriminator} $D$ with parameter $W_D$ to approximate the first term in Eq.~(\ref{elbo2}) by the following two-player game~\cite{goodfellow2014generative}:
\begin{equation}\label{gan}
\begin{split}
 \min_{{G}}\max_{{D}}\mathcal{L}_{adv}(\boldsymbol{z}, \boldsymbol{b})&=E_{\boldsymbol{o}\sim p_{\text{data}}}[D\left(\boldsymbol{o}\right)]\\&\!\!\!\!\!\!\!\!\!\!\!\!\!\!\!\!\!\!\!\!\!\!\!\!\!-\!E_{\boldsymbol{z} \sim q(\bm{z}|\bm{o}),
 \boldsymbol{b} \sim q(\bm{b}|\bm{o})}[D\left(G(\boldsymbol{z}; \theta)+\boldsymbol{b}\right)].
 \end{split}
\end{equation}

Thus, from Eqs.~(\ref{term}) and (\ref{gan}), we can reformulate the negative lower bound in Eq.~(\ref{elbo2}) as follows:
\vspace{-1mm}
\begin{equation}\label{lhat}
\begin{split}
\widehat{\mathcal{L}}\left (\boldsymbol{z},\boldsymbol{b};\boldsymbol{o} \right )= \gamma \mathcal{L}_{adv}(\boldsymbol{z}, \boldsymbol{b})&+D_{KL}\left [q\left ( \boldsymbol{z}|\boldsymbol{o}\right )||~p\left ( \boldsymbol{z} \right )\right]\\&\!\!\!\!\!\!\!\!\!\!\!\!\!\!\!\!\!\!\!\!\!\!\!\!\!\!\!\!\!\!+ D_{KL}\left [q\left ( \boldsymbol{b}|\boldsymbol{o}\right )||~p\left ( \boldsymbol{b} \right )\right],
\end{split}
\vspace{-2mm}
\end{equation}
where $\gamma$ is a hyper-parameter controlling the importance between the adversarial loss and KL divergence. The value is set empirically and will be explained in experiment section.

From the analysis above, learning the generative process of rainy image is closely related to the minimization of $\widehat{\mathcal{L}}\left (\boldsymbol{b},\boldsymbol{z};\boldsymbol{o} \right )$. For optimizing the involved network parameters $W_B$, $W_R$, $\theta$, and $W_D$,
the total objective function on the entire training dataset, can be formulated as:
\vspace{-2mm}
\begin{equation}\label{loss}
\sum\nolimits_{n=1}^{N}\widehat{\mathcal{L}}\left (\boldsymbol{z}_n,\boldsymbol{b}_n;\boldsymbol{o}_{n} \right ).
\vspace{-2mm}
\end{equation}
Note that during training, $W_B$, $W_R$, $\theta$ and $W_D$ are shared across the entire training data, leading to a general statistical distribution modelling for rainy image as well as rain layer.

Based on Eqs.~(\ref{vpr}), (\ref{vpb}), (\ref{gan}), and (\ref{lhat}), we can easily construct the inference framework as shown in Fig.~\ref{net}, called variational rain generation network (VRGNet).\footnote{More details can be found in supplementary material.}
%
\vspace{-1mm}\section{Implementation Details}\label{network}\vspace{0mm}
\small
\begin{algorithm}[t]
	\renewcommand{\algorithmicrequire}{\textbf{Input:}}
	\renewcommand{\algorithmicensure}{\textbf{Output:}}
	\caption{\small{Variational Inference for Rain Generation}}
	\label{alg1}
	\begin{algorithmic}[1]  \small
		\REQUIRE  Training data $\mathcal{D}$=$\left \{\boldsymbol{o}_{n},\boldsymbol{x}_{n}\right \}_{n=1}^{N}$, batch size $n_{b}$, $n_{\text{critic}}$ times updating of $D$ for every updating \emph{BNet}, \emph{RNet}, and \emph{G}
		\ENSURE Network parameters $W = \{W_B, W_R, \theta,W_D\}$
        \WHILE {The loss in Eq.~(\ref{loss}) is not convergent}
		\FOR{$m=1$ {\bfseries to} $n_{\text{critic}}$}
		\STATE $\{\bm{o},\bm{x}\} \leftarrow$ SampleMiniBatch($\mathcal{D},n_{b}$).
        \STATE $\{\bm{\alpha}, \bm{\beta}\} \leftarrow$ \textit{RNet}$(\bm{o};W_R)$.
        \STATE $\bm{z} \leftarrow$ \text{Reparameterization}($\bm{\alpha}, \bm{\beta}$).
        \STATE $\{\bm{\mu}, \bm{\sigma}^{2}\} \leftarrow$ \textit{BNet}$(\bm{o};W_B)$.
        \STATE $\bm{b} \leftarrow$ \text{Reparameterization}$(\bm{\mu}, \bm{\sigma}^{2})$.
        \STATE $\widehat{\bm{o}} \leftarrow$ $G(\bm{z};\theta)+\boldsymbol{b}$.
		\STATE Update $D$ with fixed \textit{BNet}, \textit{RNet}, and $G$.
		\ENDFOR
        \STATE Update \textit{BNet} with fixed \textit{RNet}, $D$, and $G$.
        \STATE Update \textit{RNet} and $G$ with fixed \textit{BNet} and $D$.
        \ENDWHILE
	\end{algorithmic}
\end{algorithm}
\normalsize

\textbf{Training Strategy.}
The entire framework in Fig.~\ref{net} is first jointly trained based on the loss function in Eq.~(\ref{loss}). The whole training procedure is summarized as Algorithm~\ref{alg1}, where we adopt the gradient penalty strategy for $D$ to stabilize the adversarial learning~\cite{gulrajani2017improved}.

After obtaining the rain generator $G$, we can use it to automatically generate sufficient rain streaks, by taking $\boldsymbol{z}$ sampled from normal distribution as the input of $G$. Based on the augmented training dataset, including original and generated pairs, we retrain current representative DL-based derainers so as to further improve their performance (see Section~\ref{removal}). It is noteworthy that the augmentation operation is implemented on the original dataset, without introducing extra training pairs requiring pre-collecting groundtruth.
\textbf{Training Details.}
During the joint training, the entire network in Fig.~\ref{net} is optimized by the Adam algorithm~\cite{Kingma2014Adam}. The initial learning rates for \emph{BNet}, \emph{RNet}, \emph{G},
and \emph{D} are $2 \times 10^{-4}, 1\times 10^{-4}, 1\times 10^{-4}$, and $4\times 10^{-4}$, respectively, and divided by 2 at epochs [400, 600, 650, 675, 690, 700]. The different initialized
learning rate settings for $G$ and $D$ are inspired by~\cite{heusel2017gans}. The prior hyper-parameter $\varepsilon_{0} ^{2}$ is set as $1\times 10^{-6}$ and the dimension $t$ of latent variable $\boldsymbol{z}$ is 128. In each epoch, the batch size $n_{b}$ is set as 18, and we randomly crop
18 $\times$ 3000 patches with size 64 $\times$ 64 pixels from the rainy image $\boldsymbol{o}$ in $\mathcal{D}$ for training. As suggested in ~\cite{gulrajani2017improved}, the
penalty coefficient in WGAN-GP is 10, and $n_{\text{critic}}$ is 5, meaning that we update $D$ 5 times for each updating of \emph{BNet}, \emph{RNet}, and \emph{G}. The coefficient $\gamma$ in Eq.~(\ref{lhat}) is empirically set as 1 for synthetic datasets and 0.01 for SPA-Data.

\begin{figure}[t]
  \centering
    \vspace{-0mm}
  \includegraphics[width=1\linewidth]{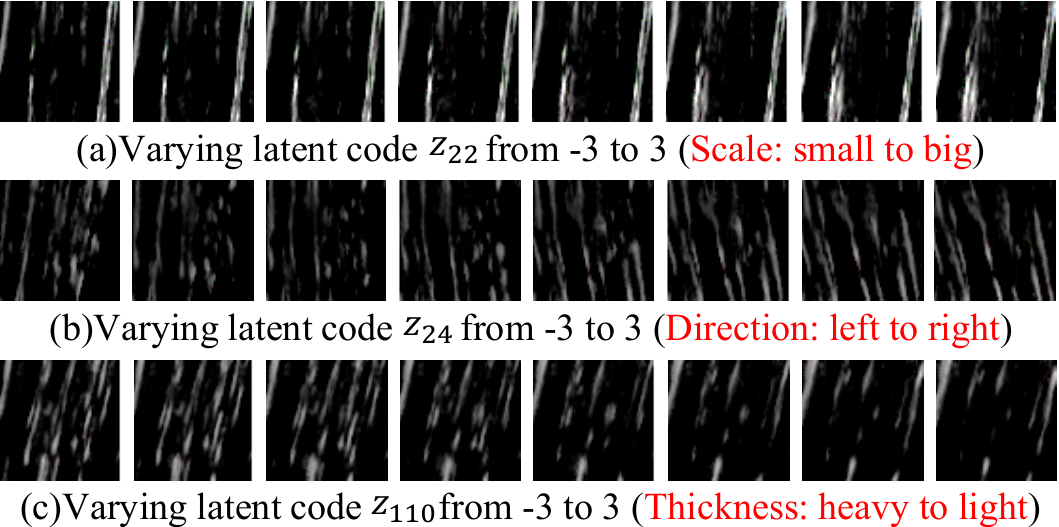}
  \vspace{-3mm}
    \caption{Manipulating latent code $\boldsymbol{z} \in \mathbb{R}^{128}$. Taking subfigure (a) as an example, we sample a random vector (latent code $\boldsymbol{z}$ ) from the normal distribution, and then only vary the latent element at the 22-th dimension of $\boldsymbol{z}$ from -3 to 3 with the interval as 0.8. Taking each varied vector $\boldsymbol{z}$ as the input of the generator $G$, the corresponding output $\boldsymbol{r}$ is each rain layer shown in (a), which demonstrates the scale property of rain. (a)-(c) denote varying different latent elements and the learned latent variables physically represent scale, direction, and thickness, respectively.}
  \label{dis}
 \vspace{-3mm}
\end{figure}
\vspace{-2mm}\section{Rain Generation Experiments}\vspace{-1mm}\label{generation}
We first conduct disentanglement and interpolation analysis to verify the potential of the VRGNet in extracting physical structural rain factors, and then evaluate the perceptual realism of our synthetic rain. Besides, with small sample experiments, we finely substantiate the effectiveness of our model in compactly capturing the manifold of rain.
\vspace{-1mm}\subsection{Disentanglement and Interpolation}\label{disen}
Similar to ~\cite{burgess2018understanding,kim2018disentangling,Chen2016InfoGAN}, we manipulate the latent code $\boldsymbol{z}$ and the disentanglement results are displayed in Fig.~\ref{dis}, where the proposed VRGNet is trained on Rain100L~\cite{Yang2019Joint}. From it, we can easily observe that these latent variables well represent interpretable physical properties in characterizing rain, including scale, direction, and thickness. Clearly, the proposed VRGNet has the capability of discovering meaningful latent rain factors, which finely complies with our latent variable modelling for rain layer in Eq.~(\ref{rsample}).

Besides, we also conduct interpolation operations in the latent space as shown in Fig.~\ref{intro} (b). The results validate that our rain generator possesses the manifold continuity in the latent space for changing the direction and thickness of rains, and thus it can generate diverse and non-repetitive rain types instead of simply memorizing the patterns in input images. More results as video clips are provided in SM.

\begin{table*}[t]
\caption{Average PSNR of PReNet on the SPA-Data test set. \textbf{Baseline} denotes that training samples are all from SPA-Data ($\sim$600K), and \textbf{GNet} means the augmented training where training samples consist of 1K real pairs randomly selected from $\sim$600K and different number of fake pairs generated by our generator that is jointly trained on $\sim$600K. \textbf{In each scene, the training pairs between Baseline and GNet keep the same}, and the result is computed over 5 random attempts. The case that \textbf{Baseline} with $\sim$600K has no randomness about samples.}\label{tabsmallsample}\vspace{1mm}
    \raggedright
    \scriptsize
     \centering
        \begin{tabular}{@{}C{3.0cm}@{}|@{}C{1.6cm}@{}|@{}C{1.6cm}@{}|@{}C{1.6cm}@{}|@{}C{1.6cm}@{}|
                    @{}C{1.6cm}@{}|@{}C{1.6cm}@{}|@{}C{1.6cm}@{}|@{}C{1.6cm}@{}|@{}C{1.2cm}@{}}
    \Xcline{1-10}{0.6pt}
{\# Real samples}                & 1K          &1.5K           &2K           &3K
                                                         & 4K          &5K           &6K           &7K
                                                         & $\sim$600K\\
    \Xcline{1-1}{0.3pt}
  { Baseline (PSNR), mean$\pm$std}                         &39.41$\pm$0.24 &39.70$\pm$0.21 &\textbf{39.86}$\pm$0.20  &39.96$\pm$0.20
                                                         &40.05$\pm$0.19 &40.04$\pm$0.18  &40.00$\pm$0.18 & 40.06$\pm$0.15
                                                         &{\textcolor{blue}{40.16}}\\
    \Xcline{1-10}{0.3pt}
    \end{tabular}
    \parbox[t][0.3mm][s]{1cm}{}
\begin{tabular}{@{}C{3.0cm}@{}|@{}C{1.6cm}@{}|@{}C{1.6cm}@{}|@{}C{1.6cm}@{}|@{}C{1.6cm}@{}|
                    @{}C{1.6cm}@{}|@{}C{1.6cm}@{}|@{}C{1.6cm}@{}|@{}C{1.6cm}@{}|@{}C{1.2cm}@{}}
   \Xcline{1-10}{0.3pt}

{\# Samples (real+fake)}                 &1K+0K            &1K+0.5K          &1K+1K             &1K+2K
                                                         &1K+3K             &1K+4K            &1K+5K             &1K+6K
                                                         & -   \\

    \Xcline{1-1}{0.3pt}
  { GNet (PSNR), mean$\pm$std}                 &39.41$\pm$0.24 &\textbf{39.71}$\pm$0.26 &39.83$\pm$0.20 &\textbf{40.25}$\pm$0.21
                                                         &\textbf{40.24}$\pm$0.17 &\textbf{40.53}$\pm$0.20 &\textbf{40.68}$\pm$0.17 & \textbf{40.70}$\pm$0.11
                                                         & - \\
    \Xcline{1-10}{0.6pt}
    \end{tabular}
    \vspace{-2mm}
\end{table*}
\begin{figure}[t]
  \centering
  \includegraphics[width=1\linewidth]{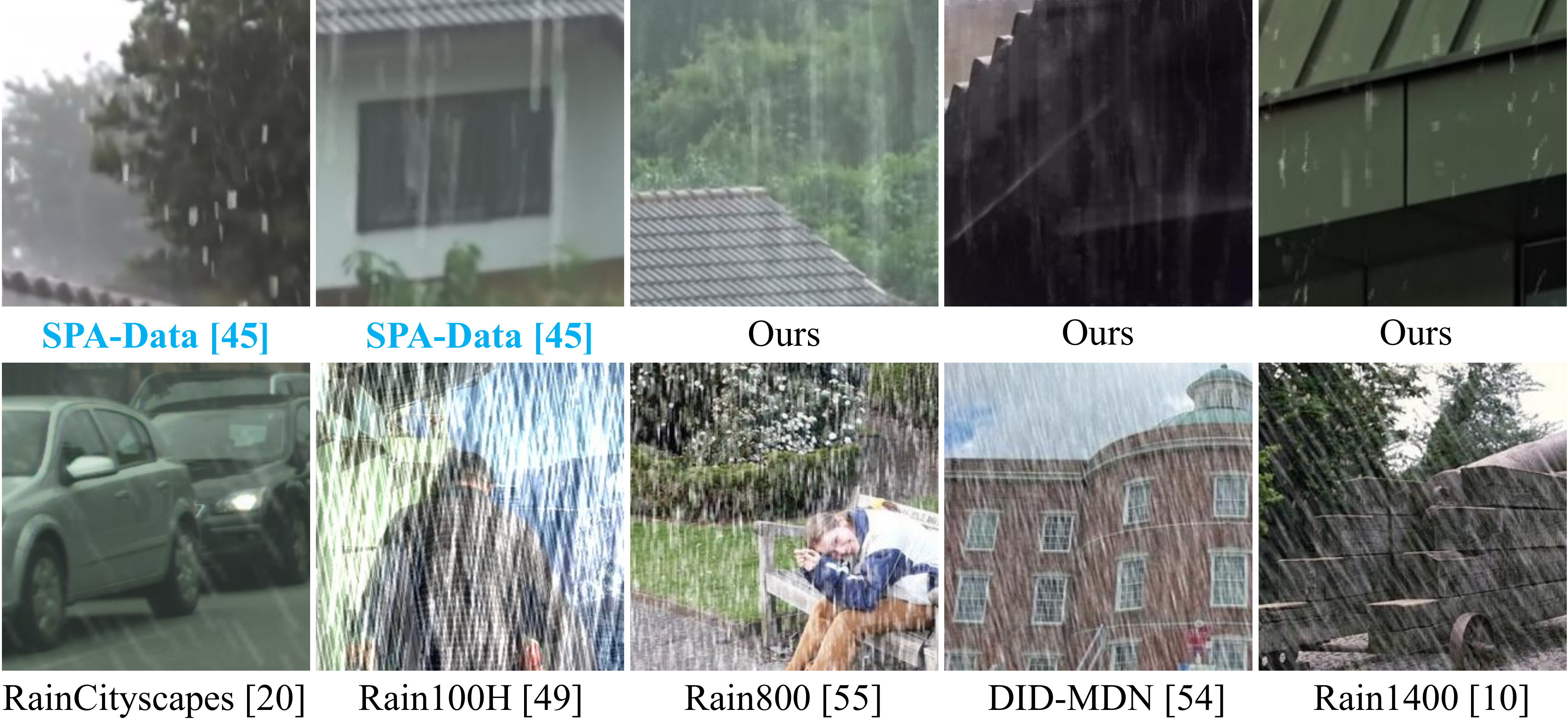}
  \vspace{-5mm}
    \caption{Rainy images randomly selected from seven different datasets. Only SPA-Data~\cite{wang2019spatial} is captured in real rain scenes.} \label{rainyimage}
  \vspace{-2mm}
\end{figure}
\begin{figure}[t]
  \centering
  \includegraphics[width=1\linewidth]{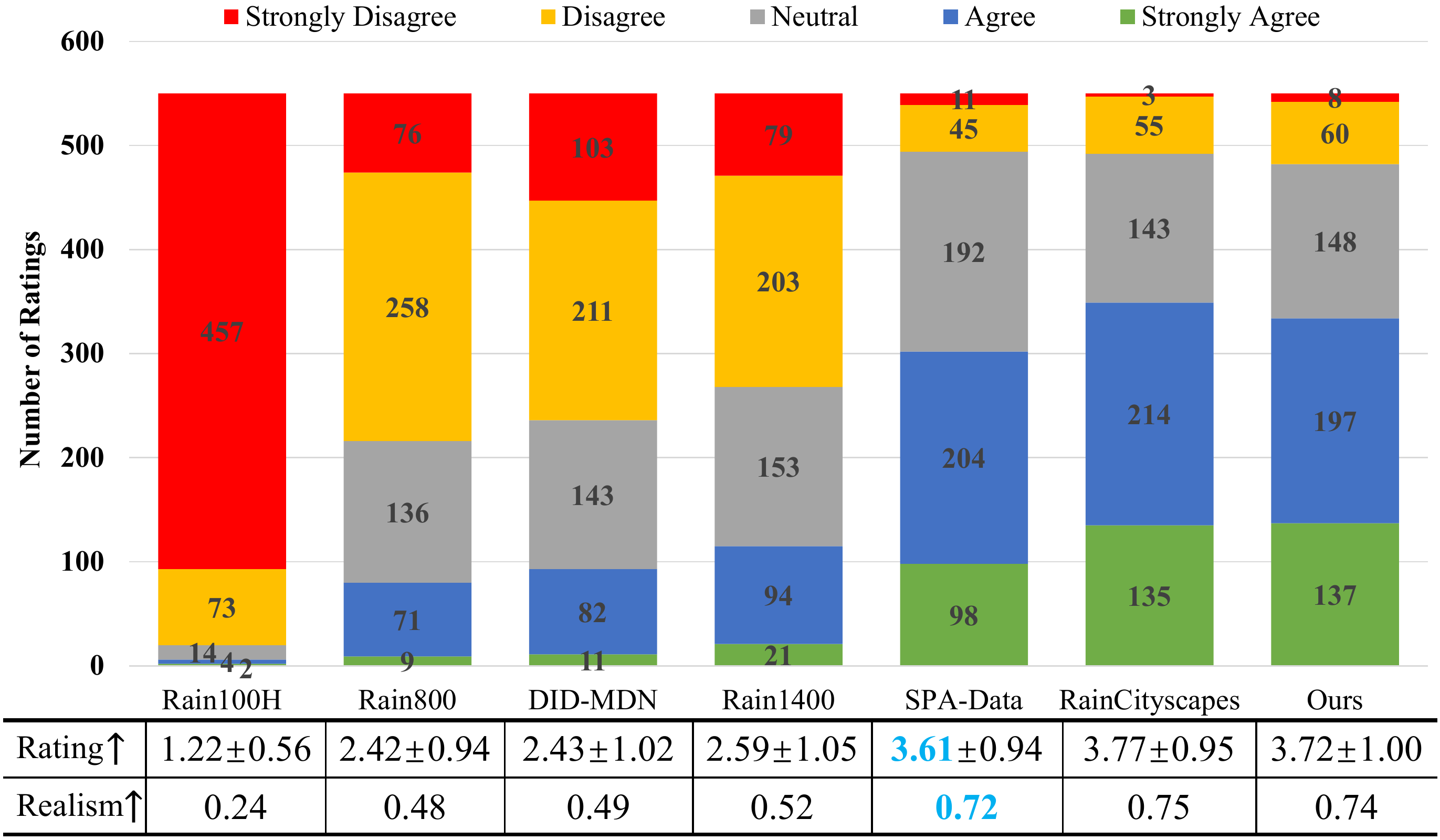}
  \vspace{-4mm}
    \caption{User study results. Upper figure: the ratings given by all participants on various datasets. Lower table: ($1^{\text{st}}$ row) the mean and standard deviation of the ratings; ($2^{\text{nd}}$ row) the realism computed by converting the mean rating to the [0,1] interval.}
  \label{rating}
 \vspace{-5mm}
\end{figure}
\vspace{-1mm}\subsection{User Study}
Fig.~\ref{rainyimage} displays the visual comparisons of rainy images randomly selected from 7 different datasets, including SPA-Data~\cite{wang2019spatial} captured in real rain scenes by controlling camera parameters, the samples randomly generated by the proposed VRGNet trained on SPA-Data, RainCityscapes~\cite{hu2019depth} generated based on a rain streak database~\cite{garg2006photorealistic}, and the other 4 synthetic datasets for rain removal, i.e., Rain1400~\cite{Fu2017Removing}, DID-MDN~\cite{zhang2018density}, Rain800~\cite{zhang2019image}, and Rain100H~\cite{Yang2019Joint}. As seen, our synthetic rains have better diversity and their appearances look closer to the real SPA-Data.

We further conduct a user study to quantitatively evaluate the quality (i.e., how realistic) of the generated rain streaks. Specifically, we prepare for 70 rainy images randomly selected from these 7 datasets with 10 samples from each dataset. Then, we recruit 55 participants with 14 females and 41 males. For each participant, we present him/her the 70 rainy images in a random order. Then they are asked to rate how real every image is, using a 5-point Likert scale. Finally, we get 550 ratings for each category.

Results are reported in Fig.~\ref{rating}, showing that our synthetic rain is judged to be significantly more realistic than most of SOTA datasets. Besides, there are three points to clarify: 1) Owning to the good diversity of generated rains (see Fig.~\ref{dis} and Fig.~\ref{rainyimage}), the ratings of our synthesized rainy images even outperform SPA-Data. 2) The realism of the synthetic RainCityscapes is slightly better than ours. However, RainCityscapes focus on modelling the fusion process of the pre-collected background layer and rain streaks that are synthesized by manually setting some model parameters~\cite{garg2006photorealistic}, and our method is for learning an interpretable generator to synthesize diverse rain streaks. From the perspective of rain layer, our method has a better capability to synthesize more diverse rain streaks than RainCityscapes (see Fig.~\ref{rainyimage}). 3) As compared with  SPA-Data and RainCityscapes, our method is able to automatically generate more sufficient and diverse rain patterns without any human intervention and empirical parameter settings, which is helpful for improving the deraining performance (see Section \ref{removal}).

As seen, our method is mainly limited by directly adopting the commonly-used addition operation in Eq.~(\ref{o}) between rain and background layer to generate rainy image. To synthesize more realistic rainy images, it is worth further exploring how to combine our rain generator and the fusion mechanism of RainCityscapes in the future.

\begin{table*}[t]
\caption{PSNR and SSIM comparisons on synthetic datasets. ``+'' denotes the augmented training. $\triangle$$\uparrow$ represents the performance gain brought by the augmented rains generated by our rain generator. \textbf{Note that the baseline of one method ``A+'' is ``A''}.}\label{tabsyn}
\centering
\vspace{-0mm}
\scriptsize
\setlength{\tabcolsep}{0.9pt}
\begin{tabular}{p{1.1cm}<{\centering}|p{0.8cm}<{\centering}|p{0.9cm}<{\centering}|p{0.9cm}<{\centering}|p{0.9cm}<{\centering}|p{0.95cm}<{\centering}p{0.95cm}<{\centering}p{0.7cm}<{\centering}|p{1cm}<{\centering}p{1.1cm}<{\centering}p{0.7cm}<{\centering} |p{1cm}<{\centering}p{1.1cm}<{\centering}p{0.7cm}<{\centering}|p{1.3cm}<{\centering}p{1.4cm}<{\centering}p{0.7cm}<{\centering}}
\Xhline{0.6pt}
\multicolumn{2}{c|}{Methods} &Input  &DSC &JCAS &DDN &DDN+ &\textcolor{blue}{$\triangle$$\uparrow$} &SPANet &SPANet+ &\textcolor{blue}{$\triangle$$\uparrow$}&PReNet &PReNet+ &\textcolor{blue}{$\triangle$$\uparrow$} &JORDER\_E &JORDER\_E+ &\textcolor{blue}{$\triangle$$\uparrow$} \\
\hline
\multirow{2}*{Rain100L}   &PSNR &26.90   &27.34 &28.54   &32.38  &35.56   &\textcolor{blue}{3.18}  &35.33  &35.83 &\textcolor{blue}{0.50}  &37.42 &37.84 &\textcolor{blue}{0.42}  &37.68  & 38.01 &\textcolor{blue}{0.33}  \\
                          &SSIM &0.838   &0.849 &0.852   &0.926  &0.966  &\textcolor{blue}{0.040}   &0.969  &0.972 &\textcolor{blue}{0.003}    &0.979 &0.980 &\textcolor{blue}{0.001}   &0.979  & 0.980 &\textcolor{blue}{0.001}   \\
\hline
\multirow{2}*{Rain100H}   &PSNR &13.56 &13.77 &14.62   &22.85  &26.99  &\textcolor{blue}{4.14}   &25.11  &27.24 &\textcolor{blue}{2.13}   &30.11  &30.48 &\textcolor{blue}{0.37}  &30.50  &32.36  &\textcolor{blue}{1.86}  \\
                          &SSIM &0.371 &0.312 &0.451   &0.725  &0.797  &\textcolor{blue}{0.072}   &0.833  &0.883 &\textcolor{blue}{0.050}  &0.905  &0.910 &\textcolor{blue}{0.005}  &0.897  &0.921 &\textcolor{blue}{0.024}\\
\hline
\multirow{2}*{Rain1400}   &PSNR &25.24  &27.88 &26.20   &28.45  &30.27 &\textcolor{blue}{1.82}   &29.85  &30.24  &\textcolor{blue}{0.39}  &32.21 &32.51 &\textcolor{blue}{0.30}   &32.00 &32.85 &\textcolor{blue}{0.85}  \\
                          &SSIM &0.810   &0.839 &0.847   &0.889  &0.917   &\textcolor{blue}{0.028}   &0.915  &0.927  &\textcolor{blue}{0.012}   &0.943 &0.945  &\textcolor{blue}{0.002}  &0.935 &0.946 &\textcolor{blue}{0.011}  \\
\Xhline{0.6pt}
\end{tabular}
\vspace{-2mm}
\end{table*}
\begin{table*}
\vspace{-0mm}
\begin{center}
\caption{Generalization performance. Average PSNR and SSIM on the test data of SPA-Data. All the DL-based methods are trained on Rain100L. The rain patterns between Rain100L and SPA-Data are quite different, which makes the generalization task hard.}
\centering
\scriptsize
\setlength{\tabcolsep}{1.1pt}
\begin{tabular}{p{1cm}<{\centering}|p{1cm}<{\centering}|p{1cm}<{\centering}|p{1cm}<{\centering}|p{1cm}<{\centering}p{1cm}<{\centering}p{0.7cm}<{\centering}|p{1cm}<{\centering}p{1.1cm}<{\centering}p{0.7cm}<{\centering}
|p{1cm}<{\centering}p{1.3cm}<{\centering}p{0.7cm}<{\centering}|p{1.4cm}<{\centering}p{1.4cm}<{\centering}p{0.7cm}<{\centering}}
\Xhline{0.6pt}
Methods &Input  &DSC &JCAS &DDN &DDN+   &\textcolor{blue}{$\triangle$$\uparrow$}  &SPANet &SPANet+  &\textcolor{blue}{$\triangle$$\uparrow$}  &PReNet &PReNet+  &\textcolor{blue}{$\triangle$$\uparrow$}  &JORDER\_E &JORDER\_E+  &\textcolor{blue}{$\triangle$$\uparrow$} \\
    \hline
   PSNR  &34.15   &34.83   &34.95  &34.66  &35.01 &\textcolor{blue}{0.35} &35.13  &35.52 &\textcolor{blue}{0.39} &34.91  &35.13 &\textcolor{blue}{0.22} &35.04  &35.15 &\textcolor{blue}{0.11} \\
 SSIM    &0.927   &0.941 &0.945  &0.935  &0.943 &\textcolor{blue}{0.008}  &0.944 &0.948 &\textcolor{blue}{0.004} &0.940  &0.942 &\textcolor{blue}{0.002}  &0.941  &0.942 &\textcolor{blue}{0.001} \\
\Xhline{0.6pt}
\end{tabular}
\label{tabltospa}
\normalsize
\end{center}
\vspace{-5mm}
\end{table*}
\begin{figure*}[!htb]
  \centering
  \includegraphics[width=1.005\linewidth]{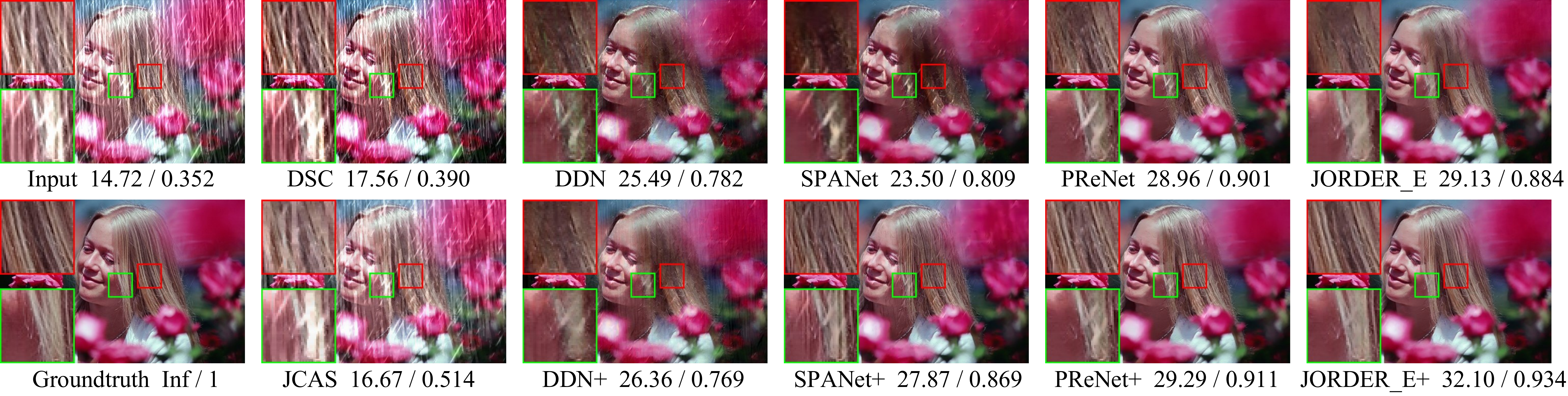}
    \vspace{-5mm}
    \caption{\textbf{Vertical contrast}. Performance comparison on a test image from Rain100H, including rainy image/groundtruth, derained results from DSC/JCAS, and deep SOTAs trained on the original ($1^{\text{st}}$ row) / augmented ($2^{\text{nd}}$ row) Rain100H training set. PSNR/SSIM is listed behind each result for easy reference. The images are better observed by zooming in on screen.}
    \label{100h}
    \vspace{-4mm}
\end{figure*}
\begin{figure*}[t]
  \centering
      \vspace{-1mm}
  \includegraphics[width=0.82\linewidth]{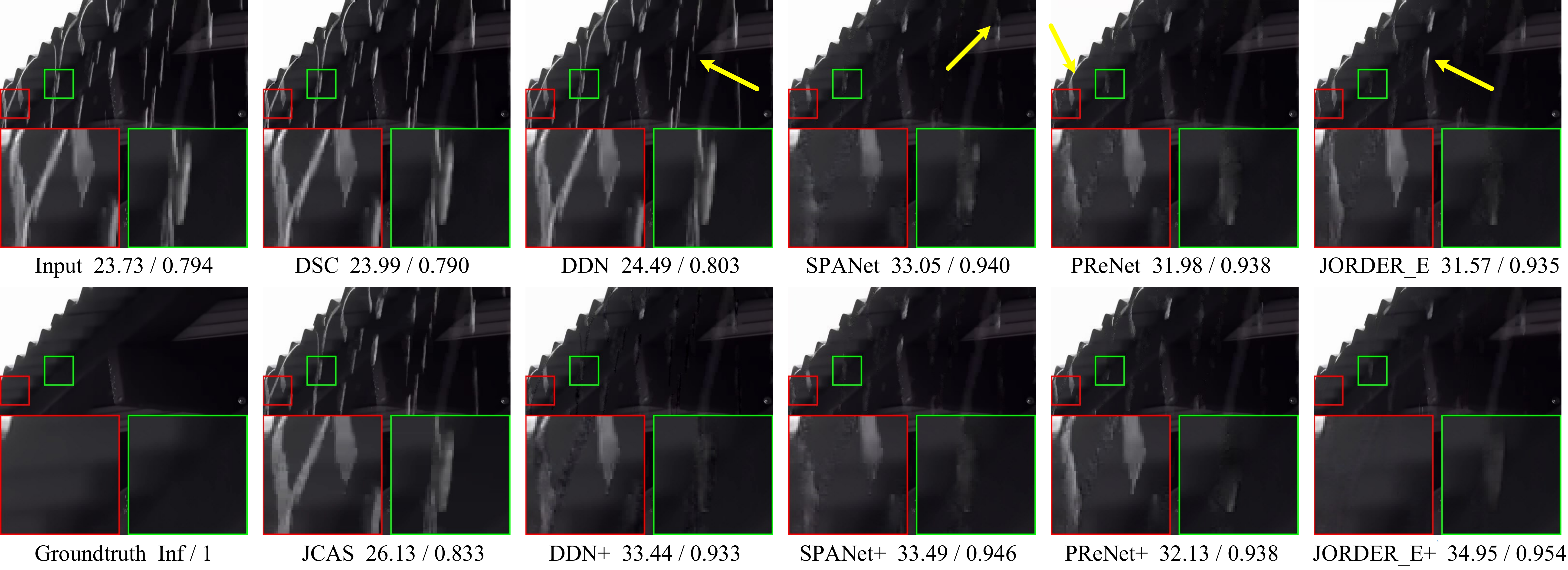}
    \vspace{-1mm}
    \caption{\textbf{Vertical contrast}. Generalization comparison on a test image from SPA-Data, including rainy image/groundtruth, derained results from DSC/JCAS, and deep derainers trained on the original ($1^{\text{st}}$ row) / augmented ($2^{\text{nd}}$ row)  Rain100L training set.}
    \label{ltospa}
    \vspace{-2mm}
\end{figure*}
\begin{figure*}[!htb]
  \centering
  \includegraphics[width=0.82\linewidth]{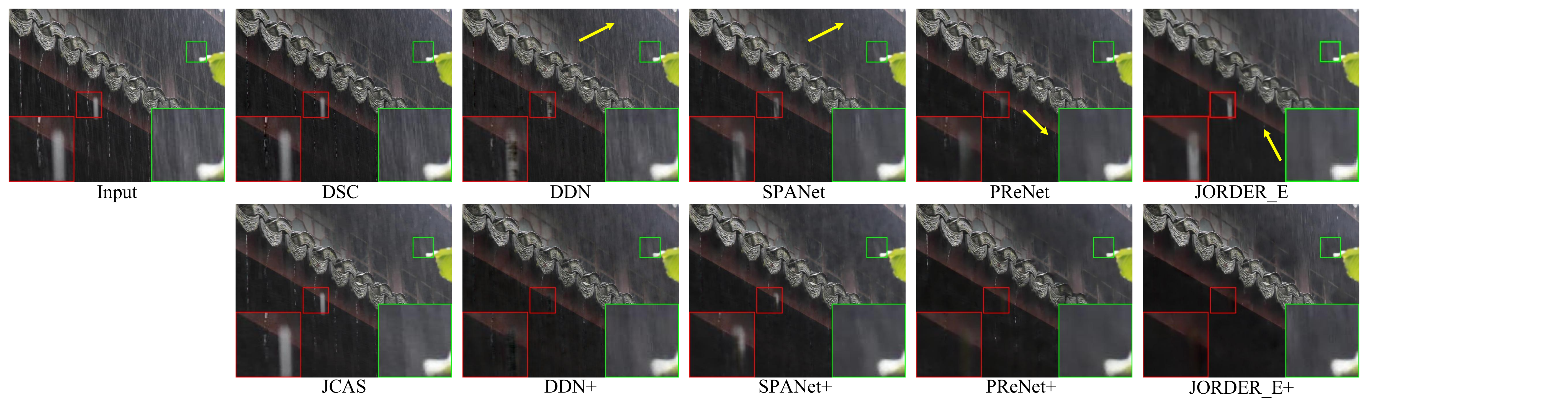}
  \vspace{-1mm}
  \caption{\textbf{Vertical contrast}. Generalization results on a real image from Internet-Data. All DL-based methods are trained on SPA-Data.}
  \vspace{-3mm}
  \label{wei}
\end{figure*}
\subsection{Small Sample Experiments on Real SPA-Data}
To further verify that our generator is able to efficiently generate more non-repetitive and diverse rain patterns, we conduct a small sample experiment on real SPA-Data with $\sim$600K training pairs and 1K test pairs. Specifically, we randomly select 1K pairs from the training set and augment them with ratio $N_{f}$ (i.e., generate $N_{f}$K fake pairs) for training. Meanwhile, we also randomly choose \textbf{the same number} (i.e., 1K+$N_{f}$K) of real pairs all from the original SPA-Data and take this case as a baseline. Due to its simplicity and fast training speed, we adopt the latest PReNet~\cite{ren2019progressive} as the deep derainer to implement this experiment.

Table~\ref{tabsmallsample} reports the PSNR averaged over 5 repetitions for different augmentation ratios.\footnote{More peak-signal-to-noise raito (PSNR)~\cite{Huynh2008Scope} and structure similarity (SSIM)~\cite{Zhou2004Image} results are listed in SM.} From it, we can observe that with the increase of ratio $N_{f}$ from 0 to 6, the average PSNR under augmented training is superior ($N_{f}$ = 2, 3, 4, 5, 6) or at least comparable ($N_{f}$ = 0, 1) to the performance (40.16~dB) under original training based on the $\sim$600K real pairs. This is mainly attributed to two points: 1) In SPA-Data, the rain scenes are not sufficiently collected to cover complicated shapes of rain streaks and many pairs are obtained by cropping one rain video shot in the same scene, which both lead to the repeatability of rain patterns. 2) The proposed generator learns the rain distribution in SPA-Data and thus can efficiently generate possible non-repetitive and diverse rain types that more compactly scatter on the manifold of such rain distribution. This also tells that the learned generator can loosen the requirement on pre-collected training samples, which is meaningful for real applications.


\vspace{-2mm}\section{Rain Removal Experiments}\label{removal}\vspace{-1mm}
Similar to the augmented strategy in ~\cite{halder2019physics}, we now utilize the generator to augment the existing datasets so as to further improve the deraining performance of current deep derainers on synthetic and real rain datasets. More experiments as well as ablation studies are included in SM.
\vspace{-1mm}\subsection{Evaluation on Synthetic Data}\vspace{-1mm}
\textbf{Representative Methods and Datasets.}
We evaluate the effectiveness of the augmentation strategy benefitted from VRGNet through latest DL-based SIRR methods, including DDN~\cite{Fu2017Removing}, PReNet~\cite{ren2019progressive},
SPANet~\cite{wang2019spatial}, and JORDER\_E~\cite{Yang2019Joint}, based on common synthetic datasets,  including Rain100L~\cite{Yang2019Joint}, Rain100H~\cite{Yang2019Joint}, and Rain1400~\cite{Fu2017Removing}. In the followings, we use notation ``A+'' to denote the results of the method A after being retrained on the augmented dataset. We also list the performance of model-based DSC~\cite{Yu2015Removing} and JCAS~\cite{Gu2017Joint} for comprehensive comparisons. To fairly compare the performance, we adopt the two commonly-used evaluation metrics, i.e., PSNR and SSIM.

\textbf{Deraining Results.}
Table~\ref{tabsyn} lists the quantitative performance of all competing methods. As seen, the deraining performance of every deep derainer after augmented training is significantly improved on all datasets and the gain $\triangle$$\uparrow$ far outperforms the sensitivity value of human visual system (about 0.1 dB). This strongly confirms that the generated rains indeed ameliorate original training sets and thus further improve the performance of DL-based methods. Naturally, the gain $\triangle$$\uparrow$ varies among different deep derainers, which is mainly caused by their different model capacities.


Fig.~\ref{100h} illustrates the visual deraining results on one hard sample from Rain100H. It is easy to observe that due to the powerful fitting capability of deep CNN, DL-based ones obviously outperform model-based DSC and JCAS. Besides, for every DL-based method, when trained on augmented dataset generated by VRGNet, its reconstructed background ($2^{\text{nd}}$ row) has better visual quality, especially in texture preservation, than the corresponding one ($1^{\text{st}}$ row) trained on original Rain100H. Clearly, the VRGNet has the potential to generate rains with higher quality and better diversity.
\vspace{-4mm}\subsection{Generalization Evaluation on Real Data}\vspace{-0mm}
We further verify the role of the generated rains in helping improve the robustness of all these deep derainers to rainy images in real-world, based on two real datasets both from~\cite{wang2019spatial}, i.e., SPA-Data and Internet-Data (no label).

\textbf{Comparisons on SPA-Data.} Table~\ref{tabltospa} quantitatively compares the generalization performance on SPA-Data where all deep derainers are trained on Rain100L. In original training, we can find that the generalization performance of all deep methods is not optimistic since the domain gap between Rain100L and SPA-Data is extremely large. Even under such a challenging scenario, after augmented training, the performance of all these methods has been improved to some extent. Note that due to larger network capacity (parameters), JORDER\_E is easier to fall into the overfitting issue and thus the performance gain $\triangle$$\uparrow$ is lower.
Fig.~\ref{ltospa} shows the derained results on a test rainy image from SPA-Data. From it, we can observe that as compared to original training, all DL-based methods with the augmented training have achieved better visual quality and higher PSNR/SSIM. Note that due to the dual influence of network structure and the quality of training set, the improvement room $\triangle$$\uparrow$ for every method is different.

\textbf{Comparisons on Internet-Data.}
Fig.~\ref{wei} shows the generalization performance on a real rainy image from Internet-Data. Under such a complex rain scene not seen in SPA-Data, these DL-based methods with augmented training evidently remove heavy rains (see the amplified red boxes). This can be rationally attributed to the diversity of generated rain types. Note that since the Internet-Data has no groundtruth, here we only provide the visual comparisons.

\vspace{-2mm}\section{Conclusion}\vspace{-1mm}
In this paper, we have explored the rain generative mechanism and constructed a full Bayesian model for generating rains from latent factors representing physical structural rain factors, such as direction, scale, and thickness. To solve this model, we have proposed a variational rain generation network (VRGNet), which implicitly infers the general statistical distribution of rains in a data-driven manner. From the learned generator, rain patches can be automatically generated to simulate diverse training samples, which facilitates a beneficial augmentation and enrichment of the existing benchmark dataset. Comprehensive rain generation verifications have fully substantiated the rationality of our generative model and evaluated the realism of the generated rain both qualitatively and quantitatively. Moreover, rain removal experiments implemented on synthetic and real datasets have finely validated the effectiveness of our generated rains in helping significantly improve the robustness of current deep single image derainers to rains in real world.

\newpage
\newpage
\newpage
\newpage
{\small
\bibliographystyle{ieee_fullname}
\bibliography{bayes_ref2}

\begin{thebibliography}{10}\itemsep=-1pt

\bibitem{blei2006variational}
David~M Blei and Michael~I Jordan.
\newblock Variational inference for {D}irichlet process mixtures.
\newblock {\em Bayesian Analysis}, 1(1):121--143, 2006.

\bibitem{burgess2018understanding}
Christopher~P Burgess, Irina Higgins, Arka Pal, Loic Matthey, Nick Watters,
  Guillaume Desjardins, and Alexander Lerchner.
\newblock Understanding disentangling in $\beta$-{V}{A}{E}.
\newblock {\em arXiv preprint arXiv:1804.03599}, 2018.

\bibitem{chen2018image}
Jingwen Chen, Jiawei Chen, Hongyang Chao, and Ming Yang.
\newblock Image blind denoising with generative adversarial network based noise
  modeling.
\newblock In {\em Proceedings of the IEEE {C}onference on {C}omputer {V}ision
  and {P}attern {R}ecognition}, pages 3155--3164, 2018.

\bibitem{Chen2016InfoGAN}
Xi Chen, Yan Duan, Rein Houthooft, John Schulman, Ilya Sutskever, and Pieter
  Abbeel.
\newblock Info{G}{A}{N}: Interpretable representation learning by information
  maximizing generative adversarial nets.
\newblock In {\em Advances in {N}eural {I}nformation {P}rocessing {S}ystems},
  2016.

\bibitem{cheng2012outdoor}
Chang Cheng, Andreas Koschan, Chung-Hao Chen, David~L Page, and Mongi~A Abidi.
\newblock Outdoor scene image segmentation based on background recognition and
  perceptual organization.
\newblock {\em IEEE Transactions on Image Processing}, 21(3):1007--1019, 2011.

\bibitem{Comaniciu2003Kernel}
Dorin Comaniciu, Visvanathan Ramesh, and Peter Meer.
\newblock Kernel-based object tracking.
\newblock {\em IEEE Transactions on Pattern Analysis and Machine Intelligence},
  25(5):564--575, 2003.

\bibitem{dinh2014nice}
Laurent Dinh, David Krueger, and Yoshua Bengio.
\newblock Nice: Non-linear independent components estimation.
\newblock {\em arXiv preprint arXiv:1410.8516}, 2014.

\bibitem{du2020variational}
Yingjun Du, Jun Xu, Qiang Qiu, Xiantong Zhen, and Lei Zhang.
\newblock Variational image deraining.
\newblock In {\em The IEEE Winter Conference on Applications of Computer
  Vision}, pages 2406--2415, 2020.

\bibitem{Fu2017Clearing}
Xueyang Fu, Jiabin Huang, Xinghao Ding, Yinghao Liao, and John Paisley.
\newblock Clearing the skies: A deep network architecture for single-image rain
  removal.
\newblock {\em IEEE {T}ransactions on {I}mage {P}rocessing}, 26(6):2944--2956,
  2017.

\bibitem{Fu2017Removing}
Xueyang Fu, Jiabin Huang, Delu Zeng, Huang Yue, Xinghao Ding, and John Paisley.
\newblock Removing rain from single images via a deep detail network.
\newblock In {\em Proceedings of the IEEE {C}onference on {C}omputer {V}ision
  and {P}attern {R}ecognition}, pages 3855--3863, 2017.

\bibitem{fu2019lightweight}
Xueyang Fu, Borong Liang, Yue Huang, Xinghao Ding, and John Paisley.
\newblock Lightweight pyramid networks for image deraining.
\newblock {\em IEEE Transactions on Neural Networks and Learning Systems},
  2019.

\bibitem{garg2006photorealistic}
Kshitiz Garg and Shree~K Nayar.
\newblock Photorealistic rendering of rain streaks.
\newblock {\em ACM Transactions on Graphics (TOG)}, 25(3):996--1002, 2006.

\bibitem{gk_vision}
Kshitiz Garg and Shree~K Nayar.
\newblock Vision and rain.
\newblock {\em {I}nternational {J}ournal of {C}omputer {V}ision}, 75(1):3--27,
  2007.

\bibitem{geiger2012are}
Andreas Geiger, Philip Lenz, and Raquel Urtasun.
\newblock Are we ready for autonomous driving? {T}he {K}{I}{T}{T}{I} vision
  benchmark suite.
\newblock pages 3354--3361, 2012.

\bibitem{goodfellow2014generative}
Ian Goodfellow, Jean Pouget-Abadie, Mehdi Mirza, Bing Xu, David Warde-Farley,
  Sherjil Ozair, Aaron Courville, and Yoshua Bengio.
\newblock Generative adversarial nets.
\newblock In {\em Advances in {N}eural {I}nformation {P}rocessing {S}ystems},
  pages 2672--2680, 2014.

\bibitem{Gu2017Joint}
Shuhang Gu, Deyu Meng, Wangmeng Zuo, and Lei Zhang.
\newblock Joint convolutional analysis and synthesis sparse representation for
  single image layer separation.
\newblock In {\em Proceedings of the IEEE {I}nternational {C}onference on
  {C}omputer {V}ision}, pages 1708--1716, 2017.

\bibitem{gulrajani2017improved}
Ishaan Gulrajani, Faruk Ahmed, Martin Arjovsky, Vincent Dumoulin, and Aaron~C
  Courville.
\newblock Improved training of wasserstein gans.
\newblock In {\em Advances in {N}eural {I}nformation {P}rocessing {S}ystems},
  pages 5767--5777, 2017.

\bibitem{halder2019physics}
Shirsendu~Sukanta Halder, Jean-Fran{\c{c}}ois Lalonde, and Raoul~de Charette.
\newblock Physics-based rendering for improving robustness to rain.
\newblock In {\em Proceedings of the IEEE {I}nternational {C}onference on
  {C}omputer {V}ision}, pages 10203--10212, 2019.

\bibitem{heusel2017gans}
Martin Heusel, Hubert Ramsauer, Thomas Unterthiner, Bernhard Nessler, and Sepp
  Hochreiter.
\newblock {GAN}s trained by a two time-scale update rule converge to a local
  nash equilibrium.
\newblock In {\em Advances in {N}eural {I}nformation {P}rocessing {S}ystems},
  pages 6626--6637, 2017.

\bibitem{hu2019depth}
Xiaowei Hu, Chi-Wing Fu, Lei Zhu, and Pheng-Ann Heng.
\newblock Depth-attentional features for single-image rain removal.
\newblock In {\em Proceedings of the IEEE Conference on Computer Vision and
  Pattern Recognition}, pages 8022--8031, 2019.

\bibitem{Huynh2008Scope}
Quan Huynh-Thu and Mohammed Ghanbari.
\newblock Scope of validity of {PSNR} in image/video quality assessment.
\newblock {\em Electronics letters}, 44(13):800--801, 2008.

\bibitem{jiang2020multi}
Kui Jiang, Zhongyuan Wang, Peng Yi, Chen Chen, Baojin Huang, Yimin Luo, Jiayi
  Ma, and Junjun Jiang.
\newblock Multi-scale progressive fusion network for single image deraining.
\newblock {\em arXiv preprint arXiv:2003.10985}, 2020.

\bibitem{jin2019unsupervised}
Xin Jin, Zhibo Chen, Jianxin Lin, Zhikai Chen, and Wei Zhou.
\newblock Unsupervised single image deraining with self-supervised constraints.
\newblock In {\em IEEE {I}nternational {C}onference on {I}mage {P}rocessing},
  pages 2761--2765. IEEE, 2019.

\bibitem{kim2019grdn}
Dong-Wook Kim, Jae Ryun~Chung, and Seung-Won Jung.
\newblock {GRDN}: Grouped residual dense network for real image denoising and
  {GAN}-based real-world noise modeling.
\newblock In {\em Proceedings of the IEEE {C}onference on {C}omputer {V}ision
  and {P}attern {R}ecognition Workshops}, pages 0--0, 2019.

\bibitem{kim2018disentangling}
Hyunjik Kim and Andriy Mnih.
\newblock Disentangling by factorising.
\newblock {\em arXiv preprint arXiv:1802.05983}, 2018.

\bibitem{Kingma2014Adam}
Diederik~P Kingma and Jimmy Ba.
\newblock Adam: A method for stochastic optimization.
\newblock {\em arXiv preprint arXiv:1412.6980}, 2014.

\bibitem{kingma2013auto}
Diederik~P Kingma and Max Welling.
\newblock Auto-encoding variational {B}ayes.
\newblock {\em arXiv preprint arXiv:1312.6114}, 2013.

\bibitem{kupyn2018deblurgan}
Orest Kupyn, Volodymyr Budzan, Mykola Mykhailych, Dmytro Mishkin, and Jiri
  Matas.
\newblock Deblur{G}{A}{N}: Blind motion deblurring using conditional
  adversarial networks.
\newblock In {\em Proceedings of the IEEE {C}onference on {C}omputer {V}ision
  and {P}attern {R}ecognition}, pages 8183--8192, 2018.

\bibitem{Li2018Non}
Guanbin Li, He Xiang, Zhang Wei, Huiyou Chang, and Lin Liang.
\newblock Non-locally enhanced encoder-decoder network for single image
  de-raining.
\newblock In {\em ACM Multimedia Conference}, 2018.

\bibitem{li2018video}
Minghan Li, Qi Xie, Qian Zhao, Wei Wei, Shuhang Gu, Jing Tao, and Deyu Meng.
\newblock Video rain streak removal by multiscale convolutional sparse coding.
\newblock In {\em Proceedings of the IEEE {C}onference on {C}omputer {V}ision
  and {P}attern {R}ecognition}, pages 6644--6653, 2018.

\bibitem{li2019heavy}
Ruoteng Li, Loongfah Cheong, and Robby~T Tan.
\newblock Heavy rain image restoration: Integrating physics model and
  conditional adversarial learning.
\newblock In {\em Proceedings of the IEEE {C}onference on {C}omputer {V}ision
  and {P}attern {R}ecognition}, pages 1633--1642, 2019.

\bibitem{Li2020All}
Ruoteng Li, Robby~T. Tan, and Loong~Fah Cheong.
\newblock All in one bad weather removal using architectural search.
\newblock In {\em IEEE/CVF {C}onference on {C}omputer {V}ision and {P}attern
  {R}ecognition}, 2020.

\bibitem{benchmark}
Siyuan Li, Iago~Breno Araujo, Wenqi Ren, Zhangyang Wang, Eric~K Tokuda,
  Roberto~Hirata Junior, Roberto Cesar-Junior, Jiawan Zhang, Xiaojie Guo, and
  Xiaochun Cao.
\newblock Single image deraining: A comprehensive benchmark analysis.
\newblock In {\em Proceedings of the IEEE {C}onference on {C}omputer {V}ision
  and {P}attern {R}ecognition}, pages 3838--3847, 2019.

\bibitem{li2018learning}
Xiaoming Li, Ming Liu, Yuting Ye, Wangmeng Zuo, Liang Lin, and Ruigang Yang.
\newblock Learning warped guidance for blind face restoration.
\newblock pages 278--296, 2018.

\bibitem{li2018recurrent}
Xia Li, Jianlong Wu, Zhouchen Lin, Hong Liu, and Hongbin Zha.
\newblock Recurrent squeeze-and-excitation context aggregation net for single
  image deraining.
\newblock In {\em Proceedings of the European Conference on Computer Vision},
  pages 254--269, 2018.

\bibitem{Li2016Rain}
Yu Li, Robby~T Tan, Xiaojie Guo, Jiangbo Lu, and Michael~S Brown.
\newblock Rain streak removal using layer priors.
\newblock In {\em Proceedings of the IEEE {C}onference on {C}omputer {V}ision
  and {P}attern {R}ecognition}, pages 2736--2744, 2016.

\bibitem{miyato2018spectral}
Takeru Miyato, Toshiki Kataoka, Masanori Koyama, and Yuichi Yoshida.
\newblock Spectral normalization for generative adversarial networks.
\newblock {\em arXiv preprint arXiv:1802.05957}, 2018.

\bibitem{Mu2019Learning}
Pan Mu, Jian Chen, Risheng Liu, Xin Fan, and Zhongxuan Luo.
\newblock Learning bilevel layer priors for single image rain streaks removal.
\newblock {\em IEEE Signal Processing Letters}, 26(2):307--311, 2019.

\bibitem{pizzati2020model}
Fabio Pizzati, Pietro Cerri, and Raoul de Charette.
\newblock Model-based occlusion disentanglement for image-to-image translation.
\newblock In {\em {E}uropean {C}onference on {C}omputer {V}ision}, 2020.

\bibitem{qian}
Rui Qian, Robby~T Tan, Wenhan Yang, Jiajun Su, and Jiaying Liu.
\newblock Attentive generative adversarial network for raindrop removal from a
  single image.
\newblock In {\em Proceedings of the IEEE {C}onference on {C}omputer {V}ision
  and {P}attern {R}ecognition}, pages 2482--2491, 2018.

\bibitem{radford2015unsupervised}
Alec Radford, Luke Metz, and Soumith Chintala.
\newblock Unsupervised representation learning with deep convolutional
  generative adversarial networks.
\newblock {\em arXiv preprint arXiv:1511.06434}, 2015.

\bibitem{ren2019progressive}
Dongwei Ren, Wangmeng Zuo, Qinghua Hu, Pengfei Zhu, and Deyu Meng.
\newblock Progressive image deraining networks: a better and simpler baseline.
\newblock In {\em Proceedings of the IEEE {C}onference on {C}omputer {V}ision
  and {P}attern {R}ecognition}, pages 3937--3946, 2019.

\bibitem{rezende2014stochastic}
Danilo~Jimenez Rezende, Shakir Mohamed, and Daan Wierstra.
\newblock Stochastic backpropagation and approximate inference in deep
  generative models.
\newblock {\em arXiv preprint arXiv:1401.4082}, 2014.

\bibitem{wang2019erl}
Guoqing Wang, Changming Sun, and Arcot Sowmya.
\newblock Erl-net: Entangled representation learning for single image
  de-raining.
\newblock In {\em Proceedings of the IEEE International Conference on Computer
  Vision}, pages 5644--5652, 2019.

\bibitem{wang2020model}
Hong Wang, Qi Xie, Qian Zhao, and Deyu Meng.
\newblock A model-driven deep neural network for single image rain removal.
\newblock In {\em Proceedings of the IEEE {C}onference on {C}omputer {V}ision
  and {P}attern {R}ecognition}, pages 3103--3112, 2020.

\bibitem{wang2019spatial}
Tianyu Wang, Xin Yang, Ke Xu, Shaozhe Chen, Qiang Zhang, and Rynson~WH Lau.
\newblock Spatial attentive single-image deraining with a high quality real
  rain dataset.
\newblock In {\em Proceedings of the IEEE {C}onference on {C}omputer {V}ision
  and {P}attern {R}ecognition}, pages 12270--12279, 2019.

\bibitem{wei2019semi}
Wei Wei, Deyu Meng, Qian Zhao, Zongben Xu, and Ying Wu.
\newblock Semi-supervised transfer learning for image rain removal.
\newblock In {\em Proceedings of the IEEE {C}onference on {C}omputer {V}ision
  and {P}attern {R}ecognition}, pages 3877--3886, 2019.

\bibitem{wei2019deraincyclegan}
Yanyan Wei, Zhao Zhang, Jicong Fan, Yang Wang, Shuicheng Yan, and Meng Wang.
\newblock Deraincyclegan: An attention-guided unsupervised benchmark for single
  image deraining and rainmaking.
\newblock {\em arXiv preprint arXiv:1912.07015}, 2019.

\bibitem{yang2017deep}
Wenhan Yang, Robby~T Tan, Jiashi Feng, Jiaying Liu, Zongming Guo, and Shuicheng
  Yan.
\newblock Deep joint rain detection and removal from a single image.
\newblock In {\em Proceedings of the IEEE {C}onference on {C}omputer {V}ision
  and {P}attern {R}ecognition}, pages 1357--1366, 2017.

\bibitem{Yang2019Joint}
Wenhan Yang, Robby~T. Tan, Jiashi Feng, Jiaying Liu, Shuicheng Yan, and
  Zongming Guo.
\newblock Joint rain detection and removal from a single image with
  contextualized deep networks.
\newblock {\em IEEE {T}ransactions on {P}attern {A}nalysis and {M}achine
  {I}ntelligence}, PP(99):1--1, 2019.

\bibitem{Yang2020Single}
Wenhan Yang, Robby~T. Tan, Shiqi Wang, Yuming Fang, and Jiaying Liu.
\newblock Single image deraining: From model-based to data-driven and beyond.
\newblock {\em IEEE {T}ransactions on {P}attern {A}nalysis \& {M}achine
  {I}ntelligence}, PP(99):1--1, 2020.

\bibitem{yasarla2019uncertainty}
Rajeev Yasarla and Vishal~M Patel.
\newblock Uncertainty guided multi-scale residual learning-using a cycle
  spinning {CNN} for single image de-raining.
\newblock In {\em Proceedings of the IEEE {C}onference on {C}omputer {V}ision
  and {P}attern {R}ecognition}, pages 8405--8414, 2019.

\bibitem{yasarla2020syn2real}
Rajeev Yasarla, Vishwanath~A Sindagi, and Vishal~M Patel.
\newblock Syn2real transfer learning for image deraining using {G}aussian
  processes.
\newblock In {\em Proceedings of the IEEE/CVF {C}onference on {C}omputer
  {V}ision and {P}attern {R}ecognition}, pages 2726--2736, 2020.

\bibitem{Yu2015Removing}
Luo Yu, Xu Yong, and Ji Hui.
\newblock Removing rain from a single image via discriminative sparse coding.
\newblock In {\em Proceedings of the IEEE {I}nternational {C}onference on
  {C}omputer {V}ision}, pages 3397--3405, 2015.

\bibitem{zhang2018self}
Han Zhang, Ian Goodfellow, Dimitris Metaxas, and Augustus Odena.
\newblock Self-attention generative adversarial networks.
\newblock {\em arXiv preprint arXiv:1805.08318}, 2018.

\bibitem{zhang2018density}
He Zhang and Vishal~M Patel.
\newblock Density-aware single image de-raining using a multi-stream dense
  network.
\newblock In {\em Proceedings of the IEEE{C}onference on {C}omputer {V}ision
  and {P}attern {R}ecognition}, pages 695--704, 2018.

\bibitem{zhang2019image}
He Zhang, Vishwanath Sindagi, and Vishal~M Patel.
\newblock Image de-raining using a conditional generative adversarial network.
\newblock {\em IEEE transactions on circuits and systems for video technology},
  2019.

\bibitem{zheng2019residual}
Yupei Zheng, Xin Yu, Miaomiao Liu, and Shunli Zhang.
\newblock Residual multiscale based single image deraining.
\newblock In {\em Conference on {B}ritish {M}achine {V}ision {C}onference},
  2019.

\bibitem{Zhou2004Image}
Wang Zhou, Bovik Alan~Conrad, Sheikh Hamid~Rahim, and Eero~P Simoncelli.
\newblock Image quality assessment: from error visibility to structural
  similarity.
\newblock {\em IEEE {T}ransactions on {I}mage {P}rocessing}, 13(4):600--612,
  2004.

\bibitem{zhu2019singe}
Hongyuan Zhu, Xi Peng, Joey~Tianyi Zhou, Songfan Yang, Vijay Chanderasekh,
  Liyuan Li, and Joo-Hwee Lim.
\newblock Single image rain removal with unpaired information: A differentiable
  programming perspective.
\newblock In {\em Proceedings of the AAAI Conference on Artificial
  Intelligence}, volume~33, pages 9332--9339, 2019.

\bibitem{zhu2017unpaired}
Jun-Yan Zhu, Taesung Park, Phillip Isola, and Alexei~A Efros.
\newblock Unpaired image-to-image translation using cycle-consistent
  adversarial networks.
\newblock In {\em Proceedings of the IEEE {I}nternational {C}onference on
  {C}omputer {V}ision}, pages 2223--2232, 2017.

\end{thebibliography}
}
\newpage

\appendix
\begin{Large}
\textbf{Supplementary Materials}
\vspace{5mm}
\end{Large}
\thispagestyle{empty}

\begin{abstract}
In this supplementary material, we provide more details on the deduction of the variational objective, and the network structures described in the main submission. Besides, we demonstrate more experimental results in rain generation and rain removal. In the end, we present ablation studies to further analyze our model.
\end{abstract}
\begin{figure*}[t]
  \centering
  \includegraphics[width=1\linewidth]{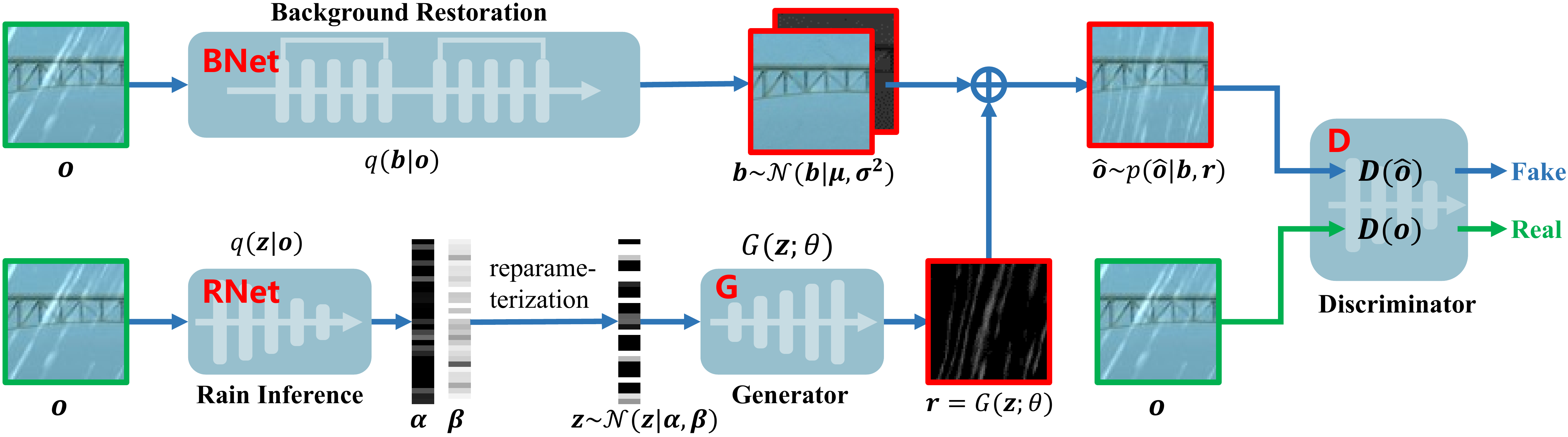}
  \vspace{-1mm}
    \caption{The flowchart of the proposed variational rain generation network (VRGNet). }
  \label{net}
   \vspace{2mm}
\end{figure*}
\section{More  Details of Variational Objective}
Here we provide a detailed derivation for variational objective in Section 3.2 of the main text.
By introducing the variational approximate posterior $q\left ( \boldsymbol{z},\boldsymbol{b}|\boldsymbol{o} \right )$, the logarithm of the rainy image distribution $p\left ( \boldsymbol{o} \right ) $ can be expressed as:
\begin{equation}\label{q1}
\begin{split}
&\textrm{log}\ p\left ( \boldsymbol{o} \right ) =\int\int q\left ( \boldsymbol{z},\boldsymbol{b}|\boldsymbol{o} \right )\textrm{log}\!\ p\left ( \boldsymbol{o} \right )\mathrm{d}\boldsymbol{z}\mathrm{d}\boldsymbol{b}\\
&= \int\int q\left ( \boldsymbol{z},\boldsymbol{b}|\boldsymbol{o} \right )\textrm{log}\left[\frac{p_\theta \left (\boldsymbol{o}|\boldsymbol{z},\boldsymbol{b} \right )p\left (\boldsymbol{z}\right )p\left (\boldsymbol{b} \right )}{p\left (\boldsymbol{z},\boldsymbol{b}|\boldsymbol{o}\right )}\right]\mathrm{d}\boldsymbol{z}\mathrm{d}\boldsymbol{b}\\
&= \int\int q\left ( \boldsymbol{z},\boldsymbol{b}|\boldsymbol{o} \right )\textrm{log}\left[\frac{p_\theta \left (\boldsymbol{o}|\boldsymbol{z},\boldsymbol{b} \right )p\left (\boldsymbol{z}\right )p\left (\boldsymbol{b} \right )}{q\left ( \boldsymbol{z},\boldsymbol{b}|\boldsymbol{o} \right )}\frac{q\left ( \boldsymbol{z},\boldsymbol{b}|\boldsymbol{o} \right )}{p\left (\boldsymbol{z},\boldsymbol{b}|\boldsymbol{o}\right )}\right]\mathrm{d}\boldsymbol{z}\mathrm{d}\boldsymbol{b}\\
&=\int\int q\left ( \boldsymbol{z},\boldsymbol{b}|\boldsymbol{o} \right )\textrm{log}\left[\frac{p_\theta \left (\boldsymbol{o}|\boldsymbol{z},\boldsymbol{b} \right )p\left (\boldsymbol{z}\right )p\left (\boldsymbol{b} \right )}{q\left ( \boldsymbol{z},\boldsymbol{b}|\boldsymbol{o} \right )}\right]\mathrm{d}\boldsymbol{z}\mathrm{d}\boldsymbol{b}\\&~~~~~~~~~~ + \int\int q\left ( \boldsymbol{z},\boldsymbol{b}|\boldsymbol{o} \right )\textrm{log}\left[\frac{q\left ( \boldsymbol{z},\boldsymbol{b}|\boldsymbol{o} \right )}{p\left (\boldsymbol{z},\boldsymbol{b}|\boldsymbol{o}\right )}\right]\mathrm{d}\boldsymbol{z}\mathrm{d}\boldsymbol{b}\\
&=E_{q\left ( \boldsymbol{z},\boldsymbol{b}|\boldsymbol{o} \right )}\!\left [\textrm{log}\!\ p_{\theta}\!\left (\boldsymbol{o}|\boldsymbol{z},\boldsymbol{b} \right )\!p\!\left ( \boldsymbol{z} \right )\! p\!\left ( \boldsymbol{b}\right ) \!-\!\textrm{log}\!\ q\!\left(\boldsymbol{z},\boldsymbol{b}|\boldsymbol{o} \right )\right ]\\&~~~~~~~~~~+D_{KL}\left [ q\left ( \boldsymbol{z},\boldsymbol{b}|\boldsymbol{o} \right ) || ~p\left (\boldsymbol{z},\boldsymbol{b}|\boldsymbol{o}\right )\right ].
\end{split}
\end{equation}
Obviously, this is just the decomposition form  given in Eq.~(7) of the main text, as:
\begin{equation}\label{hood}
\textrm{log}\!\ p\left ( \boldsymbol{o} \right ) \!=\!\mathcal{L}\left (\boldsymbol{z},\boldsymbol{b};\boldsymbol{o} \right )\!+\!D_{KL}\left [ q\left ( \boldsymbol{z},\boldsymbol{b}|\boldsymbol{o} \right ) || ~p\left (\boldsymbol{z},\boldsymbol{b}|\boldsymbol{o}\right )\right ],
\vspace{-1mm}
\end{equation}
where
\vspace{-1mm}
\begin{equation}\label{elbo}
\mathcal{L}\!\left (\boldsymbol{z},\boldsymbol{b};\boldsymbol{o} \right )\!=\!E_{q\left ( \boldsymbol{z},\boldsymbol{b}|\boldsymbol{o} \right )}\!\left [\textrm{log}\!\ p_{\theta}\!\left (\boldsymbol{o}|\boldsymbol{z},\boldsymbol{b} \right )\!p\!\left ( \boldsymbol{z} \right )\! p\!\left ( \boldsymbol{b}\right ) \!-\!\textrm{log}\!\ q\!\left(\boldsymbol{z},\boldsymbol{b}|\boldsymbol{o} \right )\right ].
\vspace{-1mm}
\end{equation}
\section{More Details on Network Architectures}
As shown in the main text, the entire network architecture is constructed as Fig.~\ref{net}, called variational rain generation network (VRGNet). It is noteworthy that we aim to propose such a variational inference framework toward rain generation without putting more emphasis on the careful design of every sub-network architecture. Specifically, each sub-network adopted in our experiment, is illustrated as:

\textit{BNet} infers posterior parameters $\boldsymbol{\mu}$ and $\boldsymbol{\sigma}^{2}$ from  $\boldsymbol{o}$ and aims to restore the latent clean background $\boldsymbol{b}$. We select the latest baseline network--PReNet~\cite{ren2019progressive} due to its simplicity and fast training process. In specific, the adopted PReNet is composed of 6 [\emph{Conv + ReLU +LSTM + ResBlocks+ Conv}] stages. The network parameters are inter-stage sharing. Besides, in each stage, the \emph{ResBlocks} consists of 5 [\emph{Conv+ReLU+Conv+ReLU+Skip connection}]  units.

\textit{RNet} helps infer the posterior parameters $\boldsymbol{\alpha}$ and $\boldsymbol{\beta}$ for latent variable $\bm{z}$, and it consists of 5 [\emph{Conv+ReLU}] blocks and a [\emph{Linear layer}] in turn.

\textit{Generator} represents the mapping $G(\boldsymbol{z}; \theta)$ for generating rain patches from extracted latent variables $\boldsymbol{z}$. Symmetrically, it contains a [\emph{Linear layer}] and 5 [\emph{Transpose Conv + ReLU}] blocks. For back propagation, we adopt the reparameterization trick as proposed in~\cite{kingma2013auto}.

\textit{Discriminator} aims to distinguish the training sample $\boldsymbol{o}$ from the generated $\widehat{\boldsymbol{o}}$, which helps the learning of $G(\boldsymbol{z}; \theta)$. Similar to the settings of most discriminators~\cite{radford2015unsupervised,zhang2018self}, the sub-network is composed of 4 [\emph{Conv + LeakyReLU}] blocks and a [Conv layer], and the negative\_slope is set as 0.1 in \emph{LeakyReLU} operation. To stabilize the training process, we also introduce the spectral normalization~\cite{miyato2018spectral} in the sub-network. Besides, motivated by ~\cite{zhang2018self}, we add the attention mechanism on the last two convolution layers to capture the global correlation in image.

Note that the number of blocks in these sub-networks, including \textit{RNet}, \textit{Generator}, and \textit{Discriminator}, is set based on the patch size (height $\times$ width of rain patches) during the network training process. In our experiments, the size is set as the commonly-used 64 $\times$ 64 in current SOTAs for this task. If other size settings are required, the number of blocks needs to be correspondingly adjusted.

\begin{figure*}[!htb]
  \centering
  \includegraphics[width=1\linewidth]{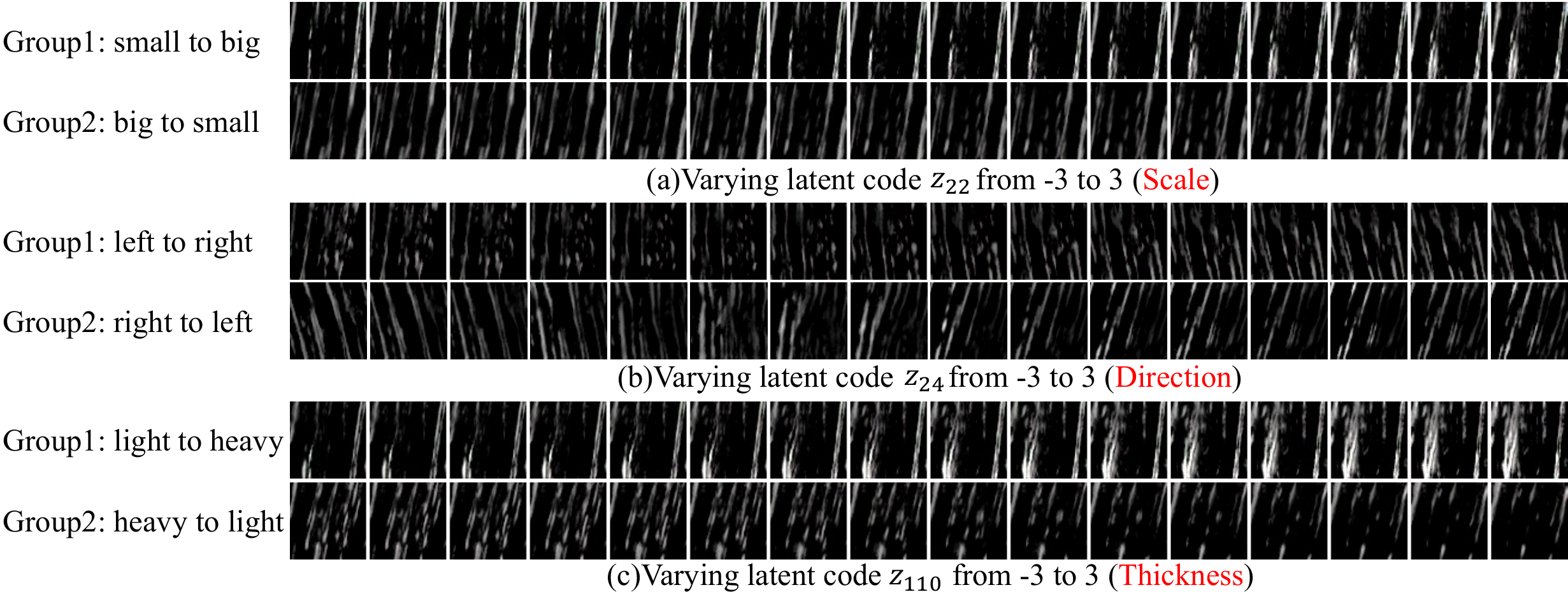}
  \vspace{-2mm}
    \caption{Manipulating latent code $\boldsymbol{z} \in \mathbb{R}^{128}$. Taking subfigure (a) as an example, we sample a random vector (latent code $\boldsymbol{z}$ ) from the normal distribution, and then only vary the latent element at the 22-th dimension of $\boldsymbol{z}$ from -3 to 3 with the interval as 0.4. Taking each varied vector $\boldsymbol{z}$ as the input of the generator $G$, the output $\boldsymbol{r}$ corresponds to each rain layer shown in (a), which demonstrates the scale property of rain. When we randomly sample two times from the normal distribution for the latent code $\boldsymbol{z}$ and repeat this experiment, the generated $\boldsymbol{r}$ are correspondingly displayed as two groups. (a)-(c) denote varying different latent elements and the learned latent variables physically represent scale, direction, and thickness, respectively.}\label{dis}
  \label{latent}
  \vspace{2mm}
\end{figure*}
\begin{figure*}[t]
  \centering
  \includegraphics[width=1\linewidth]{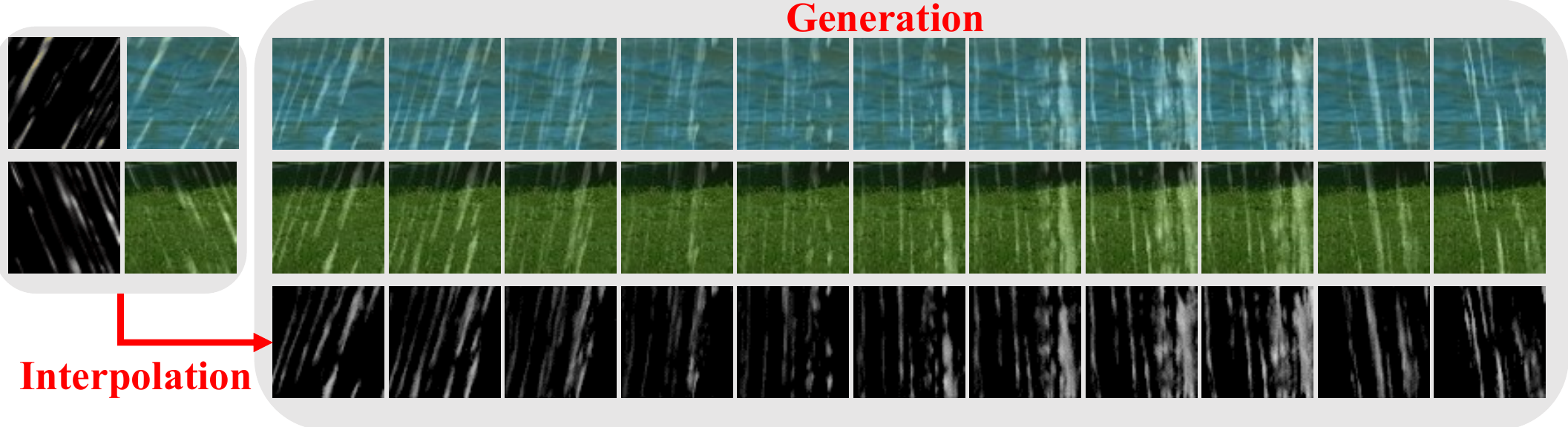}
  \vspace{-2mm}
    \caption{Interpolation. Left: two original rainy images from Rain100L and their rain layers. Right: generated rainy images (the first two rows) and synthetic rains (3rd row) obtained by linearly interpolating the latent codes of the two original rainy images.}
  \label{interpolation}
  \vspace{3mm}
\end{figure*}
\begin{table*}[t]
\caption{PSNR and SSIM (mean and standard deviation) of PReNet on the SPA-Data test set. \textbf{Baseline} denotes that training samples are all from SPA-Data ($\sim$600K), and \textbf{GNet} means the augmented training where training samples consist of 1K real pairs randomly selected from $\sim$600K and different number of fake pairs. Specifically, the fake samples are generated by our generator that is jointly trained on $\sim$600K. \textbf{In each scene, the training pairs between Baseline and GNet keep the same}, and the result is computed over 5 random attempts. The case that \textbf{Baseline} with $\sim$600K has no randomness about samples.}\label{tabsmallsample}\vspace{1mm}
    \raggedright
    \scriptsize
     \centering
        \begin{tabular}{@{}C{3.0cm}@{}|@{}C{1.6cm}@{}|@{}C{1.6cm}@{}|@{}C{1.6cm}@{}|@{}C{1.6cm}@{}|
                    @{}C{1.6cm}@{}|@{}C{1.6cm}@{}|@{}C{1.6cm}@{}|@{}C{1.6cm}@{}|@{}C{1.2cm}@{}}
    \Xcline{1-10}{0.6pt}
{\# Real samples}                & 1K          &1.5K           &2K           &3K
                                                         & 4K          &5K           &6K           &7K
                                                         & $\sim$600K\\
    \Xcline{1-1}{0.3pt}
  { Baseline (PSNR), mean$\pm$std}                         &39.41$\pm$0.24 &39.70$\pm$0.21 &\textbf{39.86}$\pm$0.20 &39.96$\pm$0.20
                                                         &40.05$\pm$0.19 &40.04$\pm$0.18  &40.00$\pm$0.18 & 40.06$\pm$0.15
                                                         &{\textcolor{blue}{40.16}}\\
    \Xcline{1-10}{0.3pt}
    \end{tabular}
    \parbox[t][0.3mm][s]{1cm}{}
\begin{tabular}{@{}C{3.0cm}@{}|@{}C{1.6cm}@{}|@{}C{1.6cm}@{}|@{}C{1.6cm}@{}|@{}C{1.6cm}@{}|
                    @{}C{1.6cm}@{}|@{}C{1.6cm}@{}|@{}C{1.6cm}@{}|@{}C{1.6cm}@{}|@{}C{1.2cm}@{}}
   \Xcline{1-10}{0.3pt}

{\# Samples (real+fake)}                 &1K+0K            &1K+0.5K          &1K+1K             &1K+2K
                                                         &1K+3K             &1K+4K            &1K+5K             &1K+6K
                                                         & -   \\

    \Xcline{1-1}{0.3pt}
  { GNet (PSNR), mean$\pm$std}                 &39.41$\pm$0.24 &\textbf{39.71}$\pm$0.26 &39.83$\pm$0.20 &\textbf{40.25}$\pm$0.21
                                                         &\textbf{40.24}$\pm$0.17 &\textbf{40.53}$\pm$0.20 &\textbf{40.68}$\pm$0.17 & \textbf{40.70}$\pm$0.11
                                                         & - \\
    \Xcline{1-10}{0.6pt}
    \end{tabular}
\parbox[t][5mm][s]{1cm}{}
        \begin{tabular}{@{}C{3.0cm}@{}|@{}C{1.6cm}@{}|@{}C{1.6cm}@{}|@{}C{1.6cm}@{}|@{}C{1.6cm}@{}|
                    @{}C{1.6cm}@{}|@{}C{1.6cm}@{}|@{}C{1.6cm}@{}|@{}C{1.6cm}@{}|@{}C{1.2cm}@{}}
    \Xcline{1-10}{0.6pt}
{\# Real samples}                & 1K          &1.5K           &2K           &3K
                                                         & 4K          &5K           &6K           &7K
                                                         & $\sim$600K\\
    \Xcline{1-1}{0.3pt}
  { Baseline (SSIM), mean$\pm$std }                         &0.9787$\pm$8e-4 &\textbf{0.9800}$\pm$7e-4 &\textbf{0.9809}$\pm$6e-4  &0.9813$\pm$7e-4
                                                         &\textbf{0.9815}$\pm$8e-4 &0.9814$\pm$5e-4  &0.9815$\pm$6e-4 & 0.9815$\pm$5e-4
                                                         &{\textcolor{blue}{0.9816}}\\
    \Xcline{1-10}{0.3pt}
    \end{tabular}
    \parbox[t][0.3mm][s]{1cm}{}
\begin{tabular}{@{}C{3.0cm}@{}|@{}C{1.6cm}@{}|@{}C{1.6cm}@{}|@{}C{1.6cm}@{}|@{}C{1.6cm}@{}|
                    @{}C{1.6cm}@{}|@{}C{1.6cm}@{}|@{}C{1.6cm}@{}|@{}C{1.6cm}@{}|@{}C{1.2cm}@{}}
   \Xcline{1-10}{0.3pt}

{\# Samples (real+fake)}                 &1K+0K            &1K+0.5K          &1K+1K             &1K+2K
                                                         &1K+3K             &1K+4K            &1K+5K             &1K+6K
                                                         & -   \\

    \Xcline{1-1}{0.3pt}
   { GNet (SSIM), mean$\pm$std }                 &0.9787$\pm$8e-4 &0.9796$\pm$6e-4 &0.9795$\pm$5e-4 &0.9813$\pm$8e-4
                                                         &0.9814$\pm$4e-4 &\textbf{0.9819}$\pm$5e-4 &\textbf{0.9820}$\pm$4e-4 & \textbf{0.9819}$\pm$4e-4
                                                         & - \\
    \Xcline{1-10}{0.6pt}
     \end{tabular}
    \vspace{2mm}
\end{table*}
\section{More Rain Generation Experiments}
In this section, we provide more disentanglement and latent space interpolation experiments to validate that the proposed rain generator is rational and can finely capture the manifold of rain underlying its implicit distribution.
\subsection{Disentanglement Experiments}
Fig.~\ref{dis} shows the resulted rain layers by manipulating the latent code $\boldsymbol{z}$ like the conventional disentanglement operations~\cite{burgess2018understanding,kim2018disentangling,Chen2016InfoGAN}. The learned generator is obtained by jointly training the proposed VRGNet based on Rain100L. From the figure, we can easily observe that these latent variables well deliver interpretable properties in generating rain layer, including direction, thickness, and scale. That is to say, the proposed VRGNet inclines to discover meaningful latent rain factors, which is finely in accordance with our modeling for rain layer by utilizing latent variables $\boldsymbol{z}$ to encode such physical structural factors.
\subsection{Latent Manifold Analysis}
We conduct interpolation operations in the latent space to estimate the manifold continuity. Here the VRGNet is jointly trained based on Rain100L. Specifically, for a pair of rainy images selected from Rain100L shown at the left of Fig.~\ref{interpolation}, we first utilize the inference model \emph{RNet} to obtain their latent codes $\boldsymbol{z}_{a}$ and $\boldsymbol{z}_{b}$, and then make linear interpolations between $\boldsymbol{z}_{a}$ and $\boldsymbol{z}_{b}$ with different weighting coefficients from 0 to 1. By inputting the weighted latent code $\boldsymbol{z}$ to the rain generator $G$, we thus synthesize different rain layers shown in the 3rd row at the right of Fig.~\ref{interpolation}. The first two rows are the generated rainy images by adding these synthetic rains on different backgrounds restored by \emph{BNet}. It is easy to observe that our rain generator has continuity in the latent space in changing the direction of rain streaks and it indeed has a fine capability to generate diverse rain types instead of simply memorizing the patterns in input images.

Note that in order to better observe the variation of rain streaks, in the interpolation experiments as shown in Fig. 1 of the main text, we have not displayed input rainy images that are used to obtain $\boldsymbol{z}$, but provided the corresponding rain layers which are easily obtained by subtracting backgrounds from the rainy images in paired testing dataset.

For better visual effect, we have conducted several groups of interpolation experiments and make each group as a file with the format `.gif' as provided in the submitted supplementary material compressed package. In these experiments, we show the variation of rain streaks in directions, thicknesses, and diversities. In each group experiment, the first and the last frames are the rain layers corresponding to one pair of input rainy images from the existing dataset, and between these two frames are the interpolated results.
\subsection{More Small Sample Experimental Results}
In this section, we also provide the SSIM results for the small sample experiments in Section 5.3 of the main text, as listed in Table~\ref{tabsmallsample}. From it, we can observe that with the increase of ratio $N_{f}$ from 0 to 6, the average PSNR and SSIM under augmented training are superior or at least comparable to the performance (40.16~dB and 0.9816) under original training based on the $\sim$600K real pairs. Please refer to the main text for more analysis.
\begin{figure*}[t]
  \centering
  \includegraphics[width=1\linewidth]{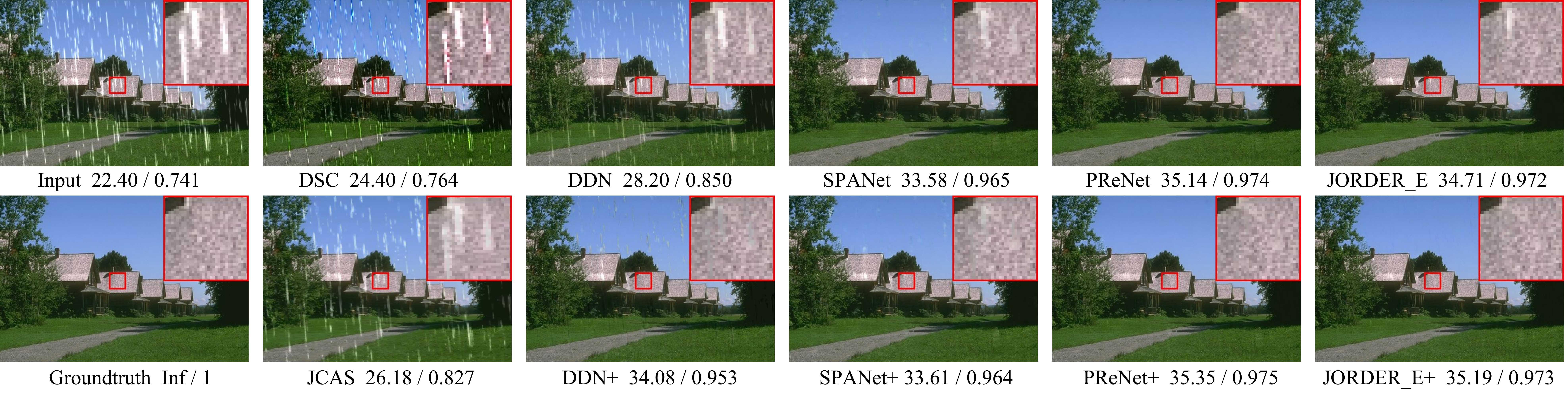}
    \vspace{-1mm}
  \caption{\textbf{Vertical contrast}. Performance comparison on a test image from Rain100L, including rainy image/groundtruth, derained results from DSC/JCAS, and deep derainers trained on the original ($1^{\text{st}}$ row) / augmented ($2^{\text{nd}}$ row) Rain100L training set. PSNR/SSIM is listed behind each result for easy reference. The images are better observed by zooming in on screen.}
  \label{100l}
 \vspace{-2mm}
\end{figure*}
\begin{figure*}[t]
  \centering
  \includegraphics[width=1\linewidth]{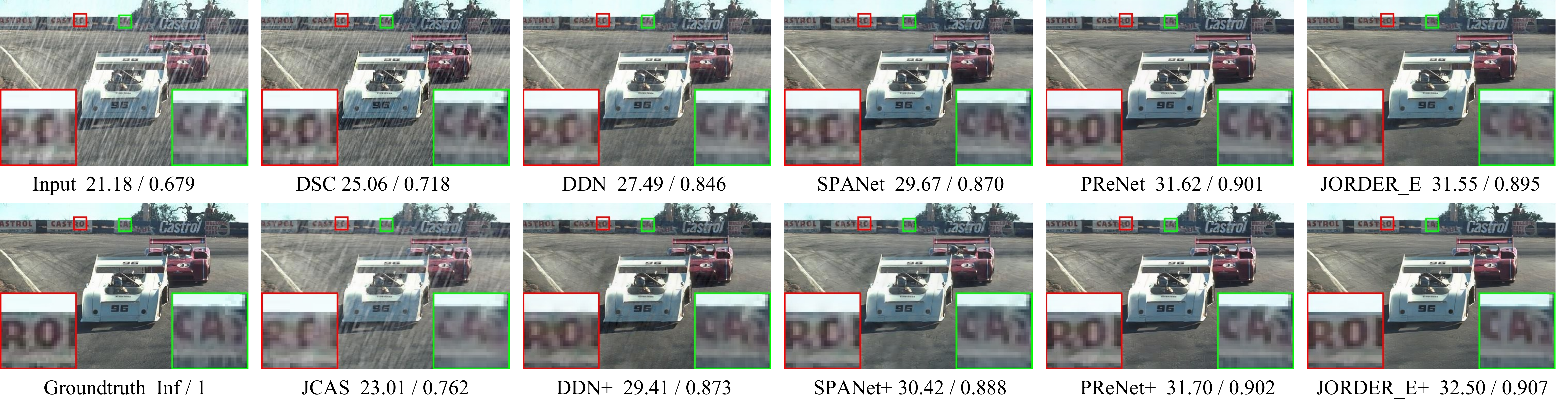}
  \vspace{-2mm}
    \caption{\textbf{Vertical contrast}. Performance comparison on a test image from Rain1400, including rainy image/groundtruth, derained results from DSC/JCAS, and deep derainers trained on the original ($1^{\text{st}}$ row) / augmented ($2^{\text{nd}}$ row) Rain1400 training set.}
  \label{1400}
   \vspace{-1mm}
\end{figure*}
\begin{table*}[!t]
\caption{PSNR and SSIM comparisons on SPA-Data testing set. ``+'' denotes the augmented training. $\triangle$$\uparrow$ represents the performance gain brought by the augmented rains generated by our rain generator that is jointly trained on the SPA-Data training set. \textbf{Note that the baseline of one method ``A+'' is ``A''}.}\label{tabspa}
\centering
\scriptsize
\setlength{\tabcolsep}{0.04pt}
\begin{tabular}{p{1.3cm}<{\centering}|p{1cm}<{\centering}|p{1cm}<{\centering}|p{1.1cm}<{\centering}|p{1.1cm}<{\centering}|p{1.1cm}<{\centering}p{1.1cm}<{\centering}p{0.6cm}<{\centering}|p{1cm}<{\centering}p{1.1cm}<{\centering}p{0.6cm}<{\centering}
|p{1.1cm}<{\centering}p{1.1cm}<{\centering}p{0.6cm}<{\centering}|p{1.4cm}<{\centering}p{1.4cm}<{\centering}p{0.6cm}<{\centering}}
\Xhline{0.6pt}
\multicolumn{2}{c|}{Methods} &Input  &DSC &JCAS &DDN &DDN+   &\textcolor{blue}{$\triangle$$\uparrow$}  &SPANet &SPANet+  &\textcolor{blue}{$\triangle$$\uparrow$}  &PReNet &PReNet+  &\textcolor{blue}{$\triangle$$\uparrow$}  &JORDER\_E &JORDER\_E+  &\textcolor{blue}{$\triangle$$\uparrow$} \\
\hline
\multirow{2}*{SPA-Data}   &PSNR &34.15   &34.95 &34.95   &36.16  &39.47 &\textcolor{blue}{3.31}  &38.14 &38.59 &\textcolor{blue}{0.45}  &40.16  &40.27 &\textcolor{blue}{0.11} &40.78 &41.49  &\textcolor{blue}{0.71}\\
                          &SSIM &0.927   &0.942 &0.945   &0.946  &0.974  &\textcolor{blue}{0.028}  &0.973  &0.974 &\textcolor{blue}{0.001} &0.981   &0.984 &\textcolor{blue}{0.003} &0.980 &0.985 &\textcolor{blue}{0.005} \\
\Xhline{0.6pt}
\end{tabular}
\vspace{2mm}
\end{table*}
\begin{figure*}[t]
  \centering
  \includegraphics[width=1\linewidth]{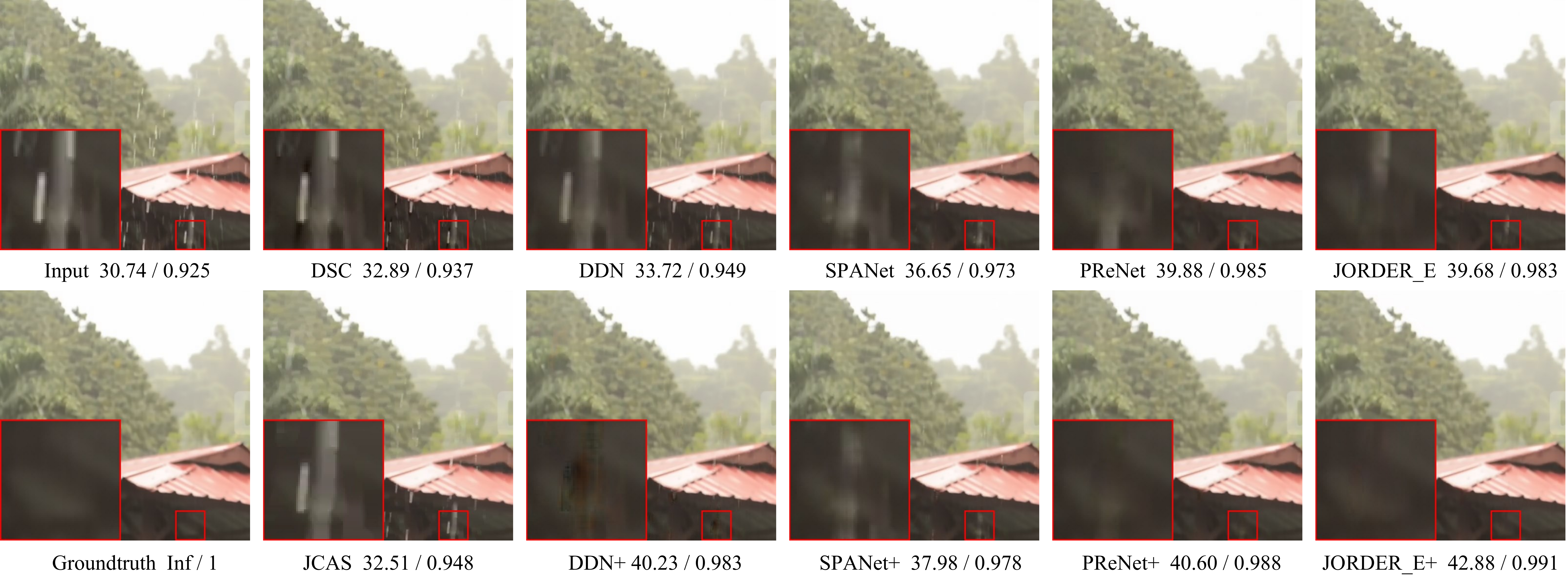}
    \vspace{-2mm}
    \caption{\textbf{Vertical contrast}. Performance comparison on a test image from SPA-Data, including rainy image/groundtruth, derained results from DSC/JCAS, and deep SOTAs trained on the original ($1^{\text{st}}$ row) / augmented ($2^{\text{nd}}$ row) SPA-Data training set.}
    \label{figspa}
    \vspace{2mm}
\end{figure*}
\begin{table}[t]
\vspace{-0mm}
\centering
\caption{PSNR and SSIM  on benchmark datasts under different cases, including jointly training the VRGNet (PReNet-) and only training BNet (PReNet).}
\footnotesize
\setlength{\tabcolsep}{3pt}
\begin{tabular}{c|cc|cc|cc|cc}
\Xhline{0.6pt}
  Datasets & \multicolumn{2}{c@{}}{Rain100L} & \multicolumn{2}{|c}{Rain100H} & \multicolumn{2}{|c@{}}{Rain1400} & \multicolumn{2}{|c@{}}{SPA-Data}\\
\Xhline{0.6pt}
  Metrics & PSNR & SSIM & PSNR & SSIM  & PSNR & SSIM & PSNR & SSIM\\
\hline
  Input & 26.90 &0.838  &13.56  &0.371    &25.24   &0.810 &34.15 &0.927\\
 PReNet-&36.94	&0.975	 &30.08	 &0.887	&32.19	&0.941	 &39.70	&0.978\\
PReNet&37.42  &0.979 &30.11 &0.905 &32.24 &0.944 &40.16 &0.981\\
\Xhline{0.6pt}
\end{tabular}
\label{tabab}
\vspace{-0mm}
\end{table}
\begin{table}[t]
\centering
\caption{Average PSNR and SSIM of PReNet on SPA-Data testing set. VRGNet- denotes the simplified VRGNet by removing \emph{BNet}, as shown in Fig.~\ref{netnobnet}. Under each setting, the result is averaged over 5 random repeated attempts.}\label{tabfake}
\scriptsize
        \begin{tabular}{@{}C{2.8cm}@{}|@{}C{1.8cm}@{}|@{}C{1.8cm}@{}|@{}C{1.8cm}@{}}
\Xhline{0.6pt}

{\# Samples (real+fake)}                     &1K+0.5K          &1K+1K             &1K+2K \\

\hline
{ VRGNet- (PSNR / SSIM) }                 &39.41 / 0.9790 &39.39 / 0.9787 &39.35 / 0.9784\\

\hline
{ VRGNet (PSNR / SSIM) }                &\textbf{39.71} / \textbf{0.9796} &\textbf{39.83} / \textbf{0.9795} &\textbf{40.25} / \textbf{0.9813}   \\
\Xhline{0.6pt}
    \end{tabular}
    \vspace{-4mm}
\end{table}
\section{More Rain Removal Experiments}
In this section, we provide more experimental results on several benchmark datasets.

\textbf{Representative Methods.}
We evaluate the effectiveness of the augmentation strategy benefitted from VRGNet through latest DL-based SIRR methods, including DDN~\cite{Fu2017Removing}, PReNet~\cite{ren2019progressive},
SPANet~\cite{wang2019spatial}, and JORDER\_E~\cite{Yang2019Joint}. In the followings, we use notation `A+' to denote the results of the method A after being retrained on the augmented dataset. Note that although our proposed VRGNet aims to help better train these DL-based SOTA derainers via data augmentation, we also list the performance of two representative model-based methods DSC~\cite{Yu2015Removing} and JCAS~\cite{Gu2017Joint} for  more comprehensive comparisons.
\subsection{More Results on Synthetic Data}
Fig.~\ref{100l} and Fig.~\ref{1400} illustrate the deraining results on two typical hard samples, from Rain100L and Rain1400, respectively. From the two figures, it is easy to observe that for every DL-based method, when trained on augmented dataset generated by VRGNet, its reconstructed background ($2^{\text{nd}}$ row) has better visual quality, especially in texture preservation, than the corresponding one ($1^{\text{st}}$ row) trained on original training set. Clearly, the VRGNet has the potential to generate rains with better diversity. Note that the performance gain varies among different deep derainers, which is mainly caused by their different model capacities.

Note that Fig.~\ref{100l} and Fig.~\ref{1400} are the performance comparisons on one test image. More quantitative comparisons on the entire testing set are listed in Table 2 of the main text.
\vspace{-0mm}\subsection{More Results on Real SPA-Data}\vspace{-0mm}
We then evaluate the effectiveness of the proposed generator on the real SPA-Data~\cite{wang2019spatial}, including $\sim$600K  training pairs and 1K testing pairs. During the augmented training phase, the exploited rain generator is trained on the entire SPA-Data training set ($\sim$600K). Note that this section represents  the same domain test experiments, instead of the generalization case as shown in Section 6.2 of the main text.

Table~\ref{tabspa} provides the quantitative results, which finely confirms the effectiveness of our proposed VRGNet in real rain generation\footnote{Note that in our all experiments, the used patch size is different from the default setting in SPANet. Under this training setting, the retrained SPANet has lower performance on SPA-Data than the original one released.}. Fig.~\ref{figspa} displays the visual comparisons on a test rainy image with complicated rain types from SPA-Data, and shows that all the DL-based derainers trained on the augmented SPA-Data have better capability in rain removal and detail recovery. Note that due to the dual influence of network structure and the quality of training set, the improvement room $\triangle$$\uparrow$ for every method is different.
\section{More Analysis about VRGNet}
\subsection{Derained Results of VRGNet}
When jointly training the VRGNet as shown in Fig.~\ref{net}, we adopt the latest PReNet~\cite{ren2019progressive} as the \emph{BNet} due to its simplicity and fast training speed.
After the joint training, the derained results (denoted as PReNet-) on benchmark datasets are reported in Table~\ref{tabab}. Naturally, we find that due to the regularization effect of adversarial loss in Eq.~(17) of the main text, the performance of PReNet- is a little lower than (but comparable to) that only training BNet (PReNet) based on the negative SSIM loss. This trend is consistent with that in most GAN based methods for low-level tasks.

%
\begin{figure*}[t]
  \centering
  \includegraphics[width=1\linewidth]{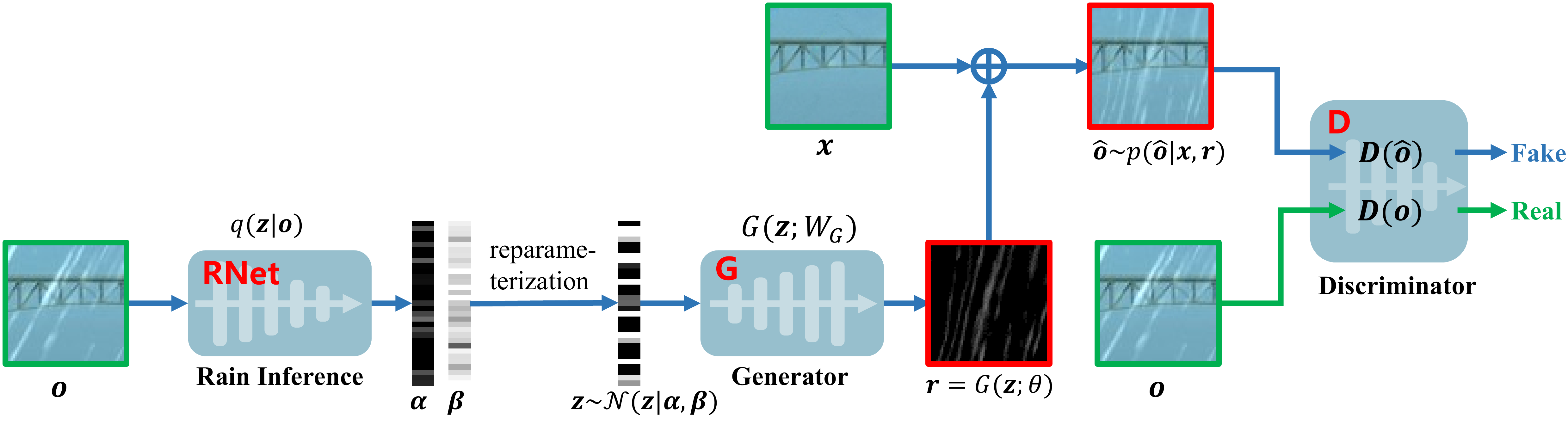}
  \vspace{-2mm}
    \caption{The flowchart of VRGNet- that directly regards $\bm{x}$ as $\bm{b}$. }
  \label{netnobnet}
\end{figure*}
\subsection{More Analysis on the Role of BNet}

From Fig.~\ref{net}, after the joint training, the \emph{BNet} does not play roles in new rain layer augmentation. However, this subnetwork is indeed necessary as analyzed below.



For convenience, we briefly denote VRGNet- as the model discarding \emph{BNet} and directly regarding the rain-free image $\bm{x}$ as the latent background $\bm{b}$, as shown in Fig.~\ref{netnobnet}. In this setting, the posterior assumption $q\left (\boldsymbol{b}|\boldsymbol{o}\right ) = \prod_{j=1}^{d}\mathcal{N}\left ( b_{j}|\mu_{j}\left (\boldsymbol{o};W_{B}\right ), \sigma _{j}^{2}\left ( \boldsymbol{o};W_{B}\right )\right )$ as Eq.~(13) of the main text can be simply set as a Dirac distribution without any parameters, i.e.,
\begin{equation}
q(\bm{b}|\bm{o}) = \text{Dirac}_{\bm{x}}(\bm{b}),
\end{equation}
where $\text{Dirac}_{\bm{x}}(\cdot)$ means the Dirac distribution centered at point $\bm{x}$. This hard assumption will lead to the degraded network framework displayed as Fig.~\ref{netnobnet}. As a special case, it indeed simplifies our proposed inference framework (Fig.~\ref{net}) to some extent, but has stricter requirements for the accuracy of the estimated rain-free image $\bm{x}$. If the pre-collected ``rain-free'' image $\bm{x}$ is not sufficiently accurate, it will naturally degrade the training performance of \emph{RNet} and the generator \emph{G}. In contrast, the introduction of \emph{BNet} is able to alleviate this issue by providing a better predicted background, and then helps  \emph{G} generate more plausible rain layer to fool discriminator \emph{D}. Therefore, we propose to adopt the more general posterior assumption as Eq.~(13) of the main text and retain \textit{BNet} in this paper.
%

To further substantiate the analysis above, we compare VRGNet- and VRGNet based on the semi-automatically generated real SPA-Data~\cite{wang2019spatial}, including $\sim$600K  training pairs and 1K testing pairs. Specifically, in SPA-Data, the rain-free image $\boldsymbol{x}$ is estimated based on multiple rainy images taken in the same condition, and thus is not the exact latent clean background $\boldsymbol{{b}}$. First, we execute the joint training on the VRGNet- (Fig.~\ref{netnobnet}) and VRGNet (Fig.~\ref{net}), respectively, based on the $\sim$600K training pairs, and obtain the corresponding different generator $G$. Then we randomly select 1K pairs from the original $\sim$600K pairs and separately augment them with ratio $N_{f}$ (i.e., generate $N_{f}$K fake pairs) by utilizing the learned two different generators.

Table~\ref{tabfake} reports the PSNR/SSIM averaged over 5 repetitions for each different augmentation ratio $N_{f}$. From the table, we can easily observe that 1) Under each $N_{f}$ setting, the performance of VRGNet significantly surpasses VRGNet-. 2) With the increase of $N_{f}$ from 0.5 to 2, the average PSNR/SSIM results of VRGNet get better while that of VRGNet- becomes worse. This is mainly because VRGNet- does not finely capture the essential rain distribution without the guidance of \emph{BNet}.

\end{document}